\pdfoutput=1
\documentclass[11pt]{article}

\usepackage{filecontents}
\usepackage{titling}
\usepackage[utf8]{inputenc}
\usepackage[english]{babel}
\usepackage{graphicx}
\usepackage{subcaption}
\usepackage{amsmath, amsfonts, amssymb, amsthm, bm, graphicx, mathtools, enumerate,multirow}
\usepackage[affil-it]{authblk}
\usepackage[round]{natbib}
\usepackage{bbm}
\usepackage{booktabs}
\usepackage{enumitem}
\usepackage{fixltx2e}
\usepackage{setspace}
\usepackage{xcolor}
\usepackage[CJKbookmarks=true,
bookmarksnumbered=true,
bookmarksopen=true,
colorlinks=true,
citecolor=blue,
linkcolor=blue,
anchorcolor=blue,
urlcolor=blue]{hyperref}
\usepackage[letterpaper, left=1.2truein, right=1.2truein, top = 1.2truein,bottom = 1.2truein]{geometry}
\usepackage[ruled, vlined, lined, commentsnumbered,linesnumbered]{algorithm2e}
\usepackage{prettyref,soul}

\setlist{nolistsep}
\allowdisplaybreaks

\usepackage{xr}
\newcommand{\E}{\textsf{E}}

\newcommand{\bY}{\bm{Y}}
\newcommand{\bJ}{\bm{J}}
\newcommand{\bA}{\bm{A}}
\newcommand{\bT}{\bm{T}}
\newcommand{\bW}{\bm{W}}
\newcommand{\bhW}{\widehat{\bW}}
\newcommand{\bAz}{\bA_{\star}}
\newcommand{\bhA}{\widehat{\bA}}

\newcommand{\bB}{\bm{B}}
\newcommand{\bC}{\bm{C}}

\newcommand{\bPi}{\bm{\Pi}}

\newcommand{\bDe}{\bm{\Delta}}

\newcommand{\bU}{\bm{U}}
\newcommand{\bu}{\bm{u}}
\newcommand{\bV}{\bm{V}}
\newcommand{\bv}{\bm{v}}

\newcommand{\norm}[1]{\Vert#1\Vert}
\newcommand{\Norm}[1]{\left\Vert#1\right\Vert}

\newcommand{\ep}[2]{#1^{\circ(#2)}}
\newcommand{\argmin}{\operatornamewithlimits{arg\ min}}
\newcommand{\tp}{\intercal}

\newcommand{\proposed}[1]{\texttt{BalWeights}\textsubscript{#1}}
\newcommand{\NW}{\texttt{NW}}
\newcommand{\KLT}{\texttt{KLT}}

\newcommand{\unitrace}{\texttt{SoftImpute}}
\newcommand{\unimax}{\texttt{CZ}}
\newcommand{\unihy}{\texttt{FLT}}

\newtheorem{lem}{Lemma}
\newtheorem{assumption}{Assumption}
\newtheorem{thm}{Theorem}

\newcommand{\bigO}{\ensuremath{\mathop{}\mathopen{}\mathcal{O}\mathopen{}}}

\newcommand{\revise}[1]{\textcolor{black}{#1}}
\newcommand{\supp}[1]{#1}

\makeatletter
\newcommand{\ltxlabel}[1]{\ltx@label{#1}}
\makeatother

\newcounter{cnt}
\newcommand{\newcnt}{
	\refstepcounter{cnt}
	\ensuremath{C_{\thecnt}}
}

\newcommand{\oldcnt}[1]{\ensuremath{C_{\ref*{#1}}}}

\begin{document}

\def\spacingset#1{\renewcommand{\baselinestretch}%
{#1}\small\normalsize}


  \title{\bf Matrix Completion with Model-free Weighting}

\author[1]{Jiayi Wang}
\author[1]{Raymond K. W. Wong}
\author[2]{Xiaojun Mao}
\author[3]{Kwun Chuen Gary Chan}
\affil[1]{Department of Statistics, Texas A\&M University}
\affil[2]{School of Data Science, Fudan University}
\affil[3]{Department of Biostatistics, University of Washington}
\date{}

\maketitle

\bigskip
\begin{abstract}
	In this paper, we propose a novel method for matrix completion under general non-uniform missing structures. By controlling an upper bound of a novel balancing error, we construct weights that can actively adjust for the non-uniformity in the empirical risk without explicitly modeling the observation probabilities, and can be computed efficiently via convex optimization. The recovered matrix based on the proposed weighted empirical risk enjoys appealing theoretical guarantees. In particular, the proposed method achieves stronger guarantee than existing work in terms of the scaling with respect to the observation probabilities, under asymptotically heterogeneous missing settings (where entry-wise observation probabilities can be of different orders). These settings can be regarded as a better theoretical model of missing patterns with highly varying probabilities. We also provide a new minimax lower bound under a class of heterogeneous settings. Numerical experiments are also provided to demonstrate the effectiveness of the proposed method.
\end{abstract}

\spacingset{1}

\section{Introduction}\label{sec:intro}
Matrix completion is a modern missing data problem
where the object of interest
is a high-dimensional and often low-rank matrix.
In its simplest form,
a partial (noisy) observation of the target matrix
is collected,
and the goal is to impute the missing entries and sometimes also to de-noise
the observed ones.
There are various related applications in, e.g.,
bioinformatics \cite{Chi-Zhou-Chen13},
causal inference \cite{athey2018matrix, Kallus-Mao-Udell18},
collaborative filtering \cite{Rennie-Srebro05}, computer vision
\cite{Weinberger-Saul06}, positioning \cite{Montanari-Oh10},
survey imputation \cite{Davenport-Plan-Berg14, Zhang-Taylor-Cobb20, Sengupta-Srebro-Evans21}
and quantum state tomography \cite{Wang13, Cai-Kim-Wang16}.
Matrix completion has been popularized by the famous Netflix prize
problem \cite{Bennett-Lanning07},
in which a large matrix of movie ratings is partially observed. Each
row of this matrix consists of ratings from a particular customer while each
column records the ratings to a particular movie.

Matrix completion has
attracted significant interest from the machine learning and
statistics communities
\citep[e.g.,][]{Koltchinskii-Lounici-Tsybakov11, Hernandez-Lobato-Houlsby-Ghahramani14, Klopp14,Lafond-Klopp-Moulines14,  Hastie-Mazumder-Lee15, Klopp-Lafond-Moulines15, Bhaskar16, Cai-Zhou16,Kang-Peng-Cheng16, Zhu-Shen-Ye16, Bi-Qu-Wang17, Fithian-Mazumder18, Dai-Wang-Shen19, Robin-Klopp-Josse20, Chen-Chi-Fan20}.
Although many statistical and computational breakthroughs \citep[e.g.,][]{Candes-Recht09,  Koltchinskii-Lounici-Tsybakov11, Recht11} have been made in this area
in the last decade,
most work (with theoretical guarantees) is developed under
a uniform missing
structure where every entry is assumed to be observed with the same probability.
However, uniform missingness is unrealistic in many applications.

The work under non-uniform missingness is relatively sparse,
and can be roughly divided into two major classes.
The first class
\citep[e.g.,][]{Srebro-Rennie-Jaakkola05, Foygel-Srebro11, Klopp14, Cai-Zhou16}
focuses on a form of robustness result, and shows
that without actively adjusting for the non-uniform missing structure
(e.g., simply applying a uniform empirical risk function $\hat{R}_{\mathrm{uni}}$ defined below),
nuclear-norm and max-norm regularized
methods can still lead to consistent estimations.
Since no direct adjustment
is imposed, there is no need to model the non-uniform missing structure.
The second class aims to improve the estimation by modeling the missing structure
and actively adjusting for non-uniformity.
\revise{Several works \citep[e.g.,][]{Srebro-Salakhutdinov10,Foygel-Shamir-Srebro11, Negahban-Wainwright12, Mao-Chen-Wong19}}
fall into this class.
\revise{However, many of the underlying models can be viewed as special low-rank (e.g., rank 1) missing structures.
For instance, a common model is the product sampling model \citep{Negahban-Wainwright12}
where row and column are chosen independently according
to possibly non-uniform marginal distributions,
leading to a rank-1 matrix of observation probability.
The specific model choices of non-uniformity
restrict the applicability and theoretical guarantees of these works.
One notable exception is \citet{Foygel-Shamir-Srebro11},
which actively adjusts for a product sampling model via a variant of weighted trace-norm regularization, but still provides guarantee under general missing structure.}
Despite these efforts, the study of non-uniform missing mechanisms
is still far from comprehensive.

In this work, we propose a novel method of \textit{balancing} weighting
to actively adjust for the non-uniform empirical risk due to general unbalanced (i.e., non-uniform) sampling,
\textit{without} explicitly modeling the probabilities of observation.
\revise{This is especially attractive when such model is hard to choose or estimate.}
We summarize our major contributions as follows.

First, we propose a novel balancing idea to adjust for the non-uniformity in matrix completion problems. Unlike many existing \revise{works}, this idea does not require specific modeling of the observation probabilities. Thanks to the proposed relaxation of the balancing error (Lemma \ref{lem:BC}), the balancing weights can then be obtained via 
a constrained spectral norm minimization, which is a convex optimization problem.

Second, we provide theoretical guarantees on the balancing performance of the proposed weights, as well as the matrix recovery via the corresponding \textit{weighted} empirical risk estimator.
We note that the estimation nature of the balancing weights introduces non-trivial dependence
in the weighted empirical risk, as opposed to the typical unweighted empirical risk (often assumed to be a sum of independent quantities).
This leads to a non-standard analysis of the proposed matrix estimator.

Third, we investigate a new type of asymptotic regime --- asymptotically heterogeneous missing structures. This regime allows observation probabilities to be of different orders, a more reasonable asymptotic model for the scenarios with highly varying probabilities among entries.
Under asymptotically heterogeneous settings,
we show that our estimator achieves a significantly better error upper bound
than existing upper bounds
in terms of the scaling with respect to the observation probabilities.
Such scaling is shown to be optimal via a new minimax result
based on a class of asymptotically heterogeneous settings.
\revise{Note that we focus on the challenging uniform error $d^2$ as opposed to the weighted (non-uniform) error $\tilde{d}^2$ (see Section \ref{sec:theory}), so as to ensure entries with high missing rate would be given non-neglible emphasis in our error measure.}

\section{Background}

\subsection{Notation}
Throughout the paper, we use several matrix norms:
nuclear norm $\|\cdot\|_*$,
Frobenius norm $\|\cdot\|_F$, spectral norm $\|\cdot\|$
entry-wise maximum norm $\|\cdot\|_\infty$ and max norm $\|\cdot\|_{\max}$.
Specifically, the \textit{entry-wise} maximum norm of a matrix $\bB=(B_{ij})$ is defined as $\|\bB\|_\infty=\max_{i,j} |B_{ij}|$,
while the max norm
is defined as
$$\|\bB\|_{\max}=\inf\{ \|\bU\|_{2,\infty}\|\bV\|_{2,\infty}: \bB=\bU\bV^\intercal\},$$
where $\|\cdot\|_{2,\infty}$ denotes the maximum $\ell_2$-row-norm of a matrix.
See, e.g., \citet{Srebro-Shraibman05} for the properties of max norm.
The Frobenius inner product and Hadamard product between two matrices
$\bB_1=(B_{1,ij})$ and $\bB_2=(B_{2,ij})$ of the same dimensions are represented by
$\langle \bB_1,\bB_2\rangle=\sum_{i,j} B_{1,ij}B_{2,ij}$ and
$\bB_1\circ\bB_2=(B_{1,ij}B_{2,ij})$ respectively.
For any $a\in\mathbb{R}$ and any matrix
$\bB=(B_{ij})$, we write
$\ep{\bB}{a}=(B_{ij}^{a})$.

We also adopt the following asymptotic notations.
Let $(b_n)_{n\ge 1}$ and $(c_n)_{n\ge 1}$ be two sequences of
nonnegative numbers.
We write $b_n = \bigO(c_n)$ if $b_n\le K c_n$ for some constant $K>0$;
and $b_n\asymp c_n$ if $b_n = \bigO(c_n)$ and $c_n=\bigO(b_n)$.
In addition, we use $\mathrm{polylog}(n)$ to represent
a polylogarithmic function of $n$, i.e., a polynomial in $\log n$.
So $\bigO(\mathrm{polylog}(n))$ represents a polylogarithmic order in $n$.

\subsection{Setup}
We aim to recover an unknown target matrix
$\bm{A}_\star=(A_{\star,ij})_{i,j=1}^{n_1,n_2}\in\mathbb{R}^{n_1\times n_2}$ from partial observation
of its noisy realization $\bY=(Y_{ij})_{i,j=1}^{n_1,n_2}\in \mathbb{R}^{n_1\times n_2}$.
Denote the observation indicator matrix $\bT=(T_{ij})_{i,j=1}^{n_1,n_2}
\in\mathbb{R}^{n_1\times n_2}$,
where $T_{ij}=1$ if
$Y_{ij}$ is observed and $T_{ij}=0$ otherwise.
We consider an additive noise model
\[
  Y_{ij}=A_{\star,ij}+\epsilon_{ij}, \quad i=1,\dots,n_1; j=1,\dots,n_2,
\]
where $\{\epsilon_{ij}\}$ are independent errors with zero mean,
and are independent of $\{T_{ij}\}$.
Also, $\{T_{ij}\}$ are independent Bernoulli random variables
with $\pi_{ij}=\Pr(T_{ij}=1)$.
We write $\bm{\Pi}=(\pi_{ij})_{i,j=1}^{n_1,n_2}$.

\subsection{Uniformity Versus Non-uniformity}
Due to complexity of data, it is often undesirable to posit an additional distributional model for
$\{\varepsilon_{ij}\}$ (such as normality) in practice.
To recover $\bA_\star$, an empirical risk minimization framework is commonly adopted
with the risk function:
\[
  R\left(\bA\right)
  =\frac{1}{n_1n_2}\E\left(\|\bY-\bA\|_{F}^2\right), \quad
  \bA\in \mathbb{R}^{n_1\times n_2}.
\]
Under uniform sampling (i.e., $\pi_{ij}\equiv\pi$),
this motivates the use of the popular empirical risk
$$\widehat{R}_{\text{uni}}(\bA) = \frac{1}{n_1 n_2} \| \bT\circ (\bY-\bA)\|_F^2,
\quad \bA\in \mathbb{R}^{n_1\times n_2},
$$
which is unbiased for $\pi R(\bA)$
\citep[e.g.,][]{Candes-Recht09,Candes-Plan10,Koltchinskii-Lounici-Tsybakov11, Klopp14}.
To minimize $\widehat{R}_{\text{uni}}$,
we can ignore the constant multiplier $\pi$.
In such settings, a popular form of estimator is
$\argmin_{\bm{A}\in\mathcal{A}_{n_1,n_2}} \widehat{R}_{\text{uni}}(\bm{A})$,
where examples of the hypothesis class $\mathcal{A}_{n_1,n_2}$
include a set of matrices with rank at most $r$
(i.e., $\{\bm{A}:\mathrm{rank}(\bm{A})\le r\}$),
and
a nuclear norm ball of radius $\nu$
(i.e., $\{\bm{A}:\|\bm{A}\|_*\le \nu\}$).
In the latter case, one can also adopt an equivalent minimization
$$\argmin_{\bm{A}}
\{\widehat{R}_{\text{uni}}(\bm{A}) + \lambda\|\bm{A}\|_*\},$$
obtained by the method of Lagrange multipliers.

However, uniform sampling is a strong assumption and
often not satisfied \citep[e.g.,][]{Srebro-Salakhutdinov10,Foygel-Shamir-Srebro11,Hernandez-Lobato-Houlsby-Ghahramani14}.
In the empirical risk minimization framework,
it is natural to adjust for such non-uniformity
since $\widehat{R}_{\text{uni}}$ is no longer unbiased for $R$.
Interestingly, such biasedness does not
lead to an incorrect estimator in an asymptotic sense
\citep{Klopp14}, a form of robustness result
(the first category of works under non-uniformity mentioned in Section \ref{sec:intro}).
This is because $\bm{A}_\star$
still minimizes $\E\{\widehat{R}_{\text{uni}}(\bm{A})\}$
even when $\pi_{ij}$'s are heterogeneous,
and, to achieve consistency, the theory \revise{requires that $\mathcal{A}_{n_1,n_2}$ grows
asymptotically} so that
some appropriate ``distance" between $\bm{A}_{\star}$
and the set $\mathcal{A}_{n_1,n_2}$
converges to zero.
For finite sample,
one often encounters some forms of misspecification ($\bm{A}_\star$ is not close to
$\mathcal{A}_{n_1,n_2}$).
In such settings,
the estimator based on $\widehat{R}_{\text{uni}}(\bm{A})$
is inclined to favor entries with a higher chance of observation,
which is often not desirable.
For movie recommendation, it is generally not a good idea to
neglect those people who rate less frequently,
as they might be the customers who do not watch as frequently,
and successful movie recommendation
would help retain these customers
from discontinuing movie subscription services.
This is highly related to misspecification in low-dimensional models
where misspecification
requires weighting adjustments \citep{Wooldridge07}.
However, matrix completion problems involve
 a much more challenging high-dimensional setup
with possibly diminishing observation probabilities \citep[e.g.,][]{Candes-Recht09, Koltchinskii11}. That is, $\pi_{L}:=\min_{i,j}\pi_{ij}\rightarrow 0$ as $n_1,n_2\rightarrow\infty$.
In fact, the diminishing setting is of great interest and plays a central role
in most analyses, since
it mimics high missing situations such as in the Netflix prize problem ($< 1\%$ of
observed ratings).

\subsection{Extremely Varying Probabilities: Heterogeneity Meets Asymptotics}

For non-uniform settings, one expects heterogeneity among the entries of $\bm{\Pi}$.
We argue that there exist different levels of heterogeneity,
and only the ``simplest" level has been well-studied.
Define
\begin{equation*}
\pi_U := \max_{i,j}\pi_{ij}\quad \mbox{and} \quad \pi_L:=\min_{i,j}\pi_{ij}.
\end{equation*}
Existing work \citep[e.g.,][]{Negahban-Wainwright12,Klopp14,Lafond-Klopp-Moulines14,Cai-Zhou16}
is based on an assumption that $\pi_U\asymp\pi_L$,
which enforces that all observation probabilities are of the same order.
We call this asymptotically homogeneous missing structure.
When the observation probabilities vary highly among different entries,
this asymptotic framework may not reflect the empirical world.
Highly varying probabilities are not rare.
As demonstrated in Section 2.3 of \citet{Mao-Wong-Chen20},
the estimated ratio of $\pi_U$ to $\pi_L$ can be high ($\ge 20000$) in
the Yahoo!~Webscope dataset, under low-rank models of $\bm\Pi$ \citep[e.g.,][]{Negahban-Wainwright12}.
In our theoretical analysis (Section \ref{sec:theory}),
we also look into the asymptotically heterogeneous settings where $\pi_U$ and $\pi_L$ are of different orders.

\section{Empirical Risk Balancing}

\subsection{Propensity Approaches and their Drawbacks}
To deal with non-uniformity,
a natural idea is
to utilize a weighted empirical risk:
\begin{equation}
  \label{eqn:Rhat}
  \widehat{R}_{\bW}\left(\bA\right)
  =\frac{1}{n_1n_2} \|{\bT\circ \ep{\bW}{1/2} \circ\left(\bY-\bA\right)}\|_{F}^2,
\end{equation}
where $\bW=(W_{ij})_{i,j=1}^{n_1,n_2}$ is a matrix composed of weights such that $W_{ij}\ge1$ for all $i$, $j$.
A natural choice of $\bm{W}$ is $(\pi_{ij}^{-1})_{i,j=1}^{n_1,n_2}$,
which leads to an unbiased risk estimator for $R(\bA)$,
and such method is
known as inverse probability weighting (IPW) in the missing data literature.
As $\{\pi_{ij}\}$ are unknown in general, most methods with IPW insert the
estimated probabilities based on certain models.
These ideas have been studied in, e.g.,
\citet{Schnabel-Swaminathan-Singh16}
under the form of a nuclear-norm regularized estimator:
\begin{equation}
\argmin_{\bA}\{\widehat{R}_{\bW}(\bm{A}) + \lambda\|\bm{A}\|_*\},
  \label{eqn:tracenorm}
\end{equation}
where $\lambda>0$ is a tuning parameter. Despite its conceptual simplicity, it is well-known
in the statistical literature
that IPW estimators could produce unstable results
due to extreme weights
\citep{Rubin01, Kang-Schafer07}.
More problematically for matrix completion, the estimation quality of a high-dimensional
probability matrix $\bm\Pi=(\pi_{ij})_{i,j=1}^{n_1,n_2}$
could also be worsened significantly by diminishing probabilities of observation
(as $n_1, n_2\rightarrow \infty$)
\citep{Davenport-Plan-Berg14}.
\revise{To solve this problem, \citet{Mao-Wong-Chen20}
imposed a constraint (effectively an upper bound) on the estimated inverse probabilities, where the constraint has to be aggressively chosen such that some true inverse probabilities do not necessarily satisfy in finite sample.}
However, there are still two general issues in this line of
research. First, the estimation of $\bm{\Pi}$
is required.
One could come up with a variety of ways to model $\bm{\Pi}$.
But it is not obvious how to choose a good model for $\bm{\Pi}$.
Second, the \revise{constraint} level is tricky to select, and difficult to analyze theoretically.
Indeed, the analysis of \revise{the effect of the constriant to matrix recovery} forms the bulk of the analysis in \citet{Mao-Wong-Chen20}.

The goal of this work is to propose a method that does not require specific
modeling and estimation of
$\bm\Pi$ but still actively adjust for the non-uniformity in the sampling. This method aims to directly find a stable
weight matrix $\bm{W}$ that adjusts for non-uniformity,
\textit{without} enforcing $\bm{W}$ to be IPW derived from a specific model.

\subsection{Balancing Weights}
\label{sec:weights}

When $\varepsilon_{ij}=0$ for all $i,j$ (only for motivation purpose, not required for the proposed techniques), we aim to choose $\bW$ such that
$\widehat{R}_{\bW}$ (left hand side) approximates the desirable ``fully-observed" one (right hand side):
\begin{equation}
  \frac{1}{n_1n_2}\|\bT\circ\ep{\bW}{1/2}\circ(\bA_\star-\bA)\|_{F}^2\approx \frac{1}{n_1n_2}\|\bA_\star-\bA\|_{F}^2,
  \label{eqn:balance1}
\end{equation}
for a set of $\bm{A}$ (a hypothesis class of $\bA_\star$ which grows with $n_1,n_2$) to be specified below.
Indeed, we only need to determine those $W_{ij}$ such that $T_{ij}=1$, since the values of the remaining $W_{ij}$ play no role in \eqref{eqn:balance1}.
Intuitively, the \textit{weights} $\bW$ are introduced to maintain \textit{balance} between the left and right hand sides of \eqref{eqn:balance1}.
Therefore, we may work with $\widehat{R}_{\bW}$ as if we were using the uniform empirical risk $\widehat{R}_{\text{uni}}$.
The condition \eqref{eqn:balance1} can be written as
\begin{align}
  0\approx 
   \frac{1}{n_1n_2}\left|\langle (\bT \circ \bW - \bm J)\circ \bDe, \bDe\rangle\right|, 
  \label{eqn:balance}
\end{align}
where $\bDe=\bA - \bA_\star$ and $\bm{J}\in\mathbb{R}^{n_1\times n_2}$ is a matrix of ones.
We call the right hand side the \textit{balancing error} of
$\bDe$ with respect to $\bW$, denoted  by $S(\bW, \bDe)$.
Naturally, we want to find weights $\bW$ that minimize the \textit{uniform} balancing error
$$F(\bW):=\sup_{\bDe\in\mathcal{D}_{n_1,n_2}} S(\bW, \bDe),$$
for a (standardized) set $\mathcal{D}_{n_1,n_2}$, induced by the hypothesis class $\mathcal{A}_{n_1,n_2}$ of $\bAz$.

A typical assumption is that $\bAz$ is low-rank or approximately low-rank.
Various classes are shown to be able to achieve such modeling.
For instance, $\mathcal{A}_{n_1,n_2}$ can be chosen as a max-norm ball
$\{\bm{A}:\|\bm{A}\|_{\max}\le \beta\}$
 \citep[e.g.,][]{Srebro-Rennie-Jaakkola05, Foygel-Srebro11,Cai-Zhou13, Cai-Zhou16, Fang-Liu-Toh18},
and the induced choice of $\mathcal{D}_{n_1,n_2}$ would be $\{\bDe: \|\bDe\|_{\max}\le 2\beta\}$.
However, the uniform balancing error does not have a closed form and
so the computation of the weights would be significantly more difficult and expensive.
Similar difficulty exists for nuclear-norm balls.

To solve this problem, we have developed the following novel lemma which allows us to focus on a
relaxed version of balancing error
that enjoys strong theoretical guarantees (see Section \ref{sec:theory}).
\begin{lem}\label{lem:BC}
  For any matrices $\bB,\bm C\in \mathbb{R}^{n_1\times n_2}$, we have
  $$|\langle \bC \circ \bB, \bB\rangle|\le \|\bC\| \|\bB\|_{\max} \|\bB\|_* \leq  \sqrt{n_1 n_2}\|\bC\| \|\bB\|_{\max}^2.$$
\end{lem}
The proof of this lemma can be found in \supp{Section \ref{appsec: BC}  of the supplemental document}.
\revise{The inequalities in Lemma 1 are tight in general:
if $\bC{ =} a\bJ$ and $\bB{ =} b\bJ$ where $a,b\in\mathbb{R}$ and $\bJ$ is the matrix whose entries are all 1, the two equalities would hold simultaneously.}

By Lemma \ref{lem:BC},
$S(\bW, \bDe) \le \sqrt{n_1 n_2}\|\bT\circ \bW - \bm J\|\|\bDe\|_{\max}^2$ 
for any $\bDe \in \mathbb{R}^{n_1\times n_2}$,
where the right hand side can be regarded as the relaxed balancing error.
If we focus on the max-norm ball (for $\mathcal{A}_{n_1,n_2}$ and hence $\mathcal{D}_{n_1,n_2}$) as discussed before,
we are only required to control the spectral norm of $\|\bT\circ\bW-\bJ\|$, which is a convex function of $\bW$.
Therefore, we propose the following novel weights:
 \begin{align}
   \label{eqn:minW}
   &\widehat{\bW}=\argmin_{\bW} \|\bT \circ \bW - \bm J\|\\
   \mbox{subject to} \quad
   &\|\bT\circ \bW\|_F\le \kappa
   \quad \mbox{and} \quad
   W_{ij}\geq 1,\nonumber
 \end{align}
where
the optimization is taken only over $W_{ij}$ such that $T_{ij}=1$.
Here $\kappa\ge \sum_{i,j}T_{ij}$ is a tuning parameter.

The weights $\{W_{ij}\}$ are restricted to be greater than or equal to 1, as their counterparts, inverse probabilities, satisfy $\pi_{ij}^{-1}\ge1$. 
The term $\|\bT\circ \bW\|_F$ regularizes $\bW$ and is particularly important
when $\varepsilon_{ij}$'s are not zero.

Let $h(\kappa)=\|\bT\circ \bhW-\bJ\|$ where $\bhW$ is defined by \eqref{eqn:minW} with the tuning parameter $\kappa$.
It is proportional to the relaxed balancing error with respect to $\bhW$.
As $\kappa$ increases, \revise{a} weaker constraint is imposed on $\bm{W}$.
Therefore $h(\kappa)$ is non-increasing as $\kappa$ increases.
It can be shown that $h(\kappa)$ stays constant for all large enough $\kappa$,
i.e., $h$ achieves its smallest value.
   The percentage of (relaxed) balancing with respect to a specific $\kappa$ is defined as $[M - h(\kappa)]/(M-m)$ where $M:= \max_\kappa h(\kappa)$ and $m=\min_{\kappa} h(\kappa)$.
    One way to tune $\kappa$ is to choose $\kappa$ that achieves certain pre-specified percentage of balancing.
    We can also select $\kappa$ from multiple values of $\kappa$
    with respect to certain balancing percentages, via a validation set.
    In Sections \ref{sec:sim} and \ref{sec:real},  we compare $\kappa$ with respect to balancing percentages $100\%$, $75\%$, and $50\%$, and select the one with the smallest validation error.

\subsection{Computation}
	The dual Lagrangian form of the constrained problem \eqref{eqn:minW} is
	\begin{equation}\label{eqn:minW2Larg}
	\underset{W_{ij}\ge 1}{\min}\left\{\|\bT\circ\bW-\bJ\|+\kappa^{\prime}\Norm{\bT\circ\bW}_{F}^2\right\},
	\end{equation}
	where $\kappa^{\prime}$ is the dual parameter.
	Denote $\bm{X}=\bT\circ\bW-\bJ$, 
	we can obtain the analytic form of the \revise{subgradient} of the largest singular value by $\partial\|\bm{X}\|=\bu_{1}^{\intercal}(\partial\bm{X})\bv_{1}$ where $\bu_1$ and $\bv_{1}$ are the corresponding left and right singular vectors with respect to the largest singular value of matrix $\bm{X}$. Thus we have
	\[
	\frac{\partial\|\bm{X}\|}{\partial W_{ij}}=\frac{\partial\|\bm{X}\|}{\partial \bm{X}}\frac{\partial\bm{X}}{\partial W_{ij}}=\bu_{1}\bv_{1}^{\intercal}T_{ij},
	\]
	and $\partial\norm{\bT\circ\bW}_{F}^2/\partial W_{ij}=2T_{ij}W_{ij}$.
  This allows us to efficiently adopt typical algorithms for smooth optimization with box-constraints such as ``L-BFGS-B" algorithm.

  \section{Estimation of \texorpdfstring{$\bAz$}{A0}}\label{sec:estA0}

  Given the weight estimator $\bhW$ defined by \eqref{eqn:minW},  we propose the following hybrid estimator that utilizes the advantages of both max-norm and nuclear-norm regularizations:
 \begin{equation}\label{eqn:objA}
  \bhA=
  \underset{\norm{\bA}_{\max}\le \beta}{\arg\min}
  \left\{ \widehat{R}_{\bhW}(\bA) + \mu\|\bA\|_* \right\},
  \end{equation}
  where $\|\cdot\|_*$ denotes the nuclear norm, and $\beta> 0$, $\mu \ge 0$ are tunning parameters.
  As explained in Section \ref{sec:weights},
  the balancing weights $\bhW$ aims to make $\widehat{R}_{\bhW}$ \revise{behave} like  the uniform empirical risk $\widehat{R}_{\mathrm{uni}}$ over a max-norm ball.
  Although not entirely necessary,
  the additional nuclear-norm penalty can sometimes produce tighter relaxation as shown in Lemma \ref{lem:BC}. As discussed in \citet{Fang-Liu-Toh18}, the additional nuclear norm bound shows its advantages under the uniform sampling scheme when the target matrix is exactly low-rank. We also find that using the hybrid of max-norm and nuclear-norm regularizations improve the estimation performance. 
  If one enforces all the elements of $\bhW$ to be 1 (uniform weighting), then the estimator \eqref{eqn:objA} degenerates to the estimator defined in \citet{Fang-Liu-Toh18}.
  The major novelty of our work is the stable weights.

  We extend the algorithm proposed in \citet{Fang-Liu-Toh18} to handle the weighted empirical risk function, so as to solve \eqref{eqn:objA}. Corresponding details can be found in \supp{Section \ref{appsec:algo_convex}  of the supplemental document}.

  \section{Theoretical Properties}
  \label{sec:theory}
  
  We provide a non-asymptotic analysis of the proposed estimator \eqref{eqn:objA}.
  One major challenge of our analysis is the estimation nature of the weights.
  As the same set of data is used to obtain the weights,
  the weighted empirical risk $\widehat{R}_{\bhW} (\bA)$ possesses complicated dependence structure, as opposed to the uniform empirical risk $\widehat{R}_{\mathrm{uni}} (\bA)$ (which is assumed to be a sum of independent variables), even for a fixed $\bA$.  \revise{To study the convergence, we carefully decompose the errors into different components. We utilize the properties of true weights  to control the balancing error term. Besides, we develop a novel lemma (Lemma \ref{applem:dualcon}) to study the concentration of the dual max-norm of the noise matrix with entry-wise multiplicative perturbation.}
  
  The following two assumptions will be used in our theoretical analysis.
  Recall that $\pi_{U}=\max_{i,j}\pi_{ij}$ and $\pi_{L}=\min_{i,j}\pi_{ij}$.
  
	  \begin{assumption}\label{ass:theta}
	  The observation indicators $\{T_{ij}\}$ are independent Bernoulli random variables
	  with $\pi_{ij}=\Pr(T_{ij}=1)$.
	  The minimum observation probability $\pi_{L} $ is positive,
	  but it can depend on $n_1$, $n_2$.
	  In particular, both $\pi_{U}$ and $\pi_{L}$ are allowed to diminish to zero when $n_1,n_2\rightarrow\infty$.
	  \end{assumption}
  
	  \begin{assumption}\label{ass:error}
		  The random errors $\{\epsilon_{ij}\}$ are independent and centered sub-Gaussian random variables
	  such that $\E(\epsilon_{ij})=0$ and $\max_{i,j}\norm{\epsilon_{ij}}_{\psi_2}\le\tau$ where
	  $\norm{\epsilon_{ij}}_{\psi_2}:=\inf\{t>0:\E[\exp(\epsilon_{ij}^2/t^2)]\le 2\}$ is the sub-Gaussian norm of $\epsilon_{ij}$. Also, $\{\epsilon_{ij}\}$ are independent of $\{T_{ij}\}$.
	  \end{assumption}

	We start with an essential result that the estimated weights $\bhW$
	possess the power to balance the non-uniform empirical risk.
	More specifically, in the following theorem,
	we derive a non-asymptotic upper bound of the uniform balancing error
	evaluated at $\bhW$,
	where the balancing error can be written as
	\[
	  S(\bm{W}, \bDe) = \frac{1}{n_1n_2} \left| \|\bT\circ \ep{\bW}{1/2} \circ \bDe\|_F^2 - \|\bDe\|_F^2 \right|.
	\]
  
	\begin{thm}\label{thm:berr}
	  Suppose Assumption \ref{ass:theta}
	  holds. Take $\kappa\ge (2\sum_{i,j} \pi^{-1}_{ij})^{1/2}$.
	  There exists an absolute constant  $\newcnt\ltxlabel{cnt:SWA}>0$ such that
	  for any $\beta'>0$,
	  \begin{align*}
	  \sup_{\|\bDe\|_{\max}\le \beta'} S(\bhW, \bDe)
		\leq \oldcnt{cnt:SWA} \frac{\beta'^2}{\sqrt{\pi_{L}(n_1\wedge n_2)}} \min\left\{ [\log(n_1+n_2)]^{1/2}, \pi_L^{-1/2}\right\},\nonumber
	  \end{align*}
	   with probability at least
	  $1-\exp\{-2^{-1}(\log 2) \pi_L^{2} \sum_{i,j} \pi_{ij}^{-1}\} - 1/(n_1+ n_2)$.
	  \end{thm}

  If $\|\bAz\|_{\max}\le \beta$, it is natural to take $\beta'=2\beta$,
  since $\|\bDe\|_{\max} =\|\bA-\bAz\|_{\max}\le 2\beta$ for any $\bA$ such that $\|\bA\|_{\max}\le \beta$.
   Therefore, we can take $\beta' = 2\beta$ in Theorem \ref{thm:berr} to achieve uniform control over the balancing error associated with the estimation \eqref{eqn:objA}.

  With the above balancing guarantee, we are now in a good position to study $\bhA$.
  Our guarantee for $\bhA$ is in terms of the uniform error
  $d^{2}(\bhA,\bAz):=(n_1n_2)^{-1}\norm{\bhA-\bAz}_{F}^2$,
  instead of the non-uniform error
  $\tilde{d}^{2}(\bhA, \bAz) = \norm{\ep{\bPi}{1/2}\circ(\bhA-\bAz)}_{F}^2/ \|\ep{\bPi}{1/2}\|_F^2$
   \citep[e.g.,][]{Klopp14,Cai-Zhou16}.
  Note that the non-uniform error $\tilde{d}^{2}(\bhA, \bAz)$ places less emphases on entries that are less likely to be observed, although the guarantee in terms of the non-uniform error can be stronger and is easier to obtain.
  In asymptotically heterogeneous missing settings
   (i.e., $\pi_U$ and $\pi_L$ are of different orders),
  entries with probabilities
  of order smaller than $\pi_U$ may be ignored within the non-uniform error in the asymptotic sense.
  Therefore it is not a good measure of performance if the guarantee over these entries are also important.
  In the following theorem, we provide a non-asymptotic error bound of our estimator \eqref{eqn:objA} (based on the estimated weights).

  \begin{thm}
	\label{thm:max}
	Suppose Assumptions \ref{ass:theta}--\ref{ass:error} hold.
	Assume $\|\bAz\|_{\max}\le \beta$, and
	$\mu = \bigO( \min\{ [\log(n_1+n_2)]^{1/2}, \pi_L^{-1/2}\} / \sqrt{\pi_{L}(n_1\wedge n_2)})$.
	Then there exists an absolute constant $\newcnt\ltxlabel{cnt:final}>0$ such that
	for any $\kappa\ge(2\sum_{i,j}\pi_{ij}^{-1})^{1/2}$,
  \begin{align*}
	d^{2}\left(\bhA,\bAz\right)
	\le \oldcnt{cnt:final} \left[\frac{\beta^2}{\sqrt{\pi_{L}(n_1\wedge n_2)}} \times\min\left\{ [\log(n_1+n_2)]^{1/2}, \pi_L^{-1/2}\right\} + \frac{\beta\tau\kappa\sqrt{n_1+n_2}}{n_1n_2}\right]\nonumber
  \end{align*}
	with probability at least $1- \exp\{-2^{-1}(\log 2) \pi_L^{2} \sum_{i,j} \pi_{ij}^{-1}\} - 2\exp\{-(n_1+n_2)\} - 1/(n_1 + n_2)$.
  \end{thm}

  First, we consider
  the asymptotically homogeneous missing structures (i.e., $\pi_L \asymp \pi_U$)
  which most existing work assumes.
  Under $\pi_L\asymp \pi_U$,
  the two errors $d^2(\bhA, \bAz)$ and $\tilde{d}^2(\bhA, \bAz)$
  are of the same order because
  \begin{equation}
	\label{eqn:two_d}
	\frac{\pi_L}{\pi_U}d^2(\bhA, \bAz) \le \tilde{d}^2(\bhA, \bAz)\le \frac{\pi_U}{\pi_L} d^2(\bhA, \bAz).
  \end{equation}
  \revise{Therefore, the upper bound for $\tilde{d}^2(\bhA, \bAz)$ that most existing work provides can be directly used to derive an upper bound
  for ${d}^2(\bhA, \bAz)$, which shares the same order.
  Note that $\pi_U$ and $\pi_L$ are allowed to be different despite $\pi_U\asymp \pi_L$.
  So certain non-uniform missing structures are still allowed under the setting of asymptotically homogeneous missingness.
  This setting has been studied in \citet{Negahban-Wainwright12,Klopp14,Lafond-Klopp-Moulines14,Cai-Zhou16}.
  Our bound is directly comparable to the work of \citet{Cai-Zhou16}
  which studies a max-norm constrained estimation.
  Their result assumes $\|\bAz\|_\infty\le \alpha$ for some $\alpha$,
  which allows their bound to depend on $\alpha \beta$ instead of $\beta^2$
  as in our bound.
  The comparision of error bounds between
  max-norm-constrained estimation and nuclear-norm-regularized estimation
  is given
  in Section 3.5 of \citet{Cai-Zhou16}.
  As for exactly low-rank matrices,
  we can further show that our estimator achieves optimal error bound (up to a logarithmic order).
  Roughly speaking, if $\kappa$ is small (so weights are close to constant),
  our estimator would behave
  like a standard nuclear-norm regularized estimator, and hence share the (near-)optimality of such estimator.
  We provide the error bound of our estimator under exactly low-rank setting and asymptotically homogeneous missingness, in Theorem \ref{thm:uni_lr} of the supplemental document.}

  For non-uniform missing structures,
  the orders of $\pi_U$ and $\pi_L$ do not necessarily match.
  When their orders are different, we call these missing structures asymptotically heterogeneous.
  We now focus on how the upper bound depends on $\pi_U$ and $\pi_L$.
  As mentioned before, existing results are scarce.
  Recently, \citet{Mao-Wong-Chen20} (their Section 5.3)
  provided an extension of existing upper bounds to possibly asymptotically heterogeneous settings, with a careful analysis.
  Corresponding upper bound scales with $\pi_L^{-1}{\pi_U^{1/2}}$.
  They also provided an additional result when one has access to the \textit{true}
  probabilities $\bPi$, and show that the upper bound of the estimator based on the empirical risk defined via the true probabilities can achieve the scaling $\pi_L^{-1/2}$, which is significantly better than $\pi_L^{-1}{\pi_U^{1/2}}$.
  However, until now, it remains unclear whether there exists an estimator
  with this scaling of $\pi_U$ and $\pi_L$, without access to the true probabilities.
  Interestingly, Theorem \ref{thm:max} provides a positive result, and 
  shows that the upper bound for the proposed estimator achieves this scaling $\pi_L^{-1/2}$ under very mild assumption that $\pi_L$ is diminishing in at least a slow order, more specifically $\pi_L = \bigO(1/\log(n_1+n_2))$.
  
  Next, we provide a theoretical result indicating that
  the scaling $\pi_L^{-1/2}$ cannot be improved
  under the asymptotically heterogeneous missing structures.
  In below, we give a minimax lower bound based on a class of asymptotically heterogeneous settings.
  To the best of the authors' knowledge, the minimax lower bounds under asymptotically heterogeneous regimes have never been studied.

  The heterogeneous class that we consider posits
  \begin{equation}
	(n_1n_2)^{-1}\sum_{i=1}^{n_1}\sum_{j=1}^{n_2}\pi_{ij}\asymp \pi_{L}.
	\label{eqn:minmax_miss}
  \end{equation}
  It is clear that \eqref{eqn:minmax_miss} does not exclude asymptotically homogeneous settings.
  To demonstrate the heterogeneity, we provide an example as follows.
  Suppose there is only a fixed number of entries with observation probabilities in constant order, and the observation probabilities of the remaining entries are of the same order as $\pi_L$. Then $\pi_U\asymp 1$, and \eqref{eqn:minmax_miss} is satisfied.
  Therefore, for any diminishing $\pi_L$, this setting is asymptotically heterogeneous.
  
  Now, we provide the minimax result.
  \begin{thm}\label{thm:unilowerbound}
	Let $\{\epsilon_{ij}\}$ be \text{i.i.d.} Gaussian $\mathcal{N}(0,\sigma^2)$ with $\sigma^2>0$. For any $\beta>0$,
	assume \eqref{eqn:minmax_miss} holds with
	 \revise{\(\pi_L^{-1} = \bigO(\beta^2 (n_1\wedge n_2)/(\sigma\wedge \beta)^2)\)}.
	  Then, there exist 
	 constants $\delta\in(0,1)$ and $c>0$ such that
	  \[
	  \underset{\bhA}{\inf}\underset{\Norm{\bAz}_{\max}\le\beta}{\sup}\Pr\left(d^{2}\left(\bhA,\bAz\right)>\frac{c(\sigma \wedge \beta)\beta}{\sqrt{\pi_L\left(n_1\wedge n_2\right)}}\right)\ge\delta.
	\]
  \end{thm}
  
  \revise{In the discussion below, we focus on $\sigma\asymp 1$, which, most notably, excludes asymptotically noiseless settings.}
  Theorem \ref{thm:unilowerbound} shows that the scaling $\pi_L^{-1/2}$
  in our upper bound obtained in Theorem \ref{thm:max} is essential.
  \revise{Due to
  the general inequality
  \citep{Srebro-Shraibman05}:
  \begin{equation}\label{eqn:max_base_ineq}
	\|\bAz\|_\infty \le \|\bAz\|_{\max}\le \sqrt{\mathrm{rank}(\bAz)}\|\bAz\|_{\infty},
  \end{equation}
  $\beta$ is not expected to grow fast for low-rank $\bAz$ with bounded entries.
  For $\beta=\bigO(\mathrm{polylog}(n))$, our upper bound matches with the lower
  bound in Theorem \ref{thm:unilowerbound} up to a logarithmic factor.
  For general $\beta$, our upper bound scales with $\beta^2$ instead of $(\sigma \wedge \beta)\beta$ despite its
  matching scaling with respect to $\pi_L$.
  Indeed, a mismatch between the upper bound and the lower bound also occurs
   in \citet{Cai-Zhou16} under asymptotically homogeneous settings,
   where their bound is derived via an additional assumption $\|\bAz\|_\infty\le \alpha$.
   Their upper bound scales with $\alpha\beta$
   instead of $(\sigma\wedge \alpha)\beta$ as in their minimax lower bound.
   We leave a more detailed study of the scaling with respect to $\beta$ as a future direction.}

   \section{\revise{Simulations}}
   \label{sec:sim}
	In this simulation study, we let the target matrix $\bAz \in \mathbb{R}^{n_1 \times n_2}$ be generated by $\bAz = \bm U \bm V^\tp$, where $\bm U \in \mathbb{R}^{n_1 \times r}, \bm V \in \mathbb{R}^{n_2 \times r}$,  and each entry  of $\bm U$ and $\bm V$ is sampled uniformly and independently from $[0,2]$. We set $n_1 = n_2 = 200$ and $r=5$. Therefore, the rank of the target matrix is 5. The contaminated version of $\bAz$ is then generated as $\bm Y = \bAz + \bm \epsilon$, where $\bm \epsilon \in \mathbb{R}^{n_1 \times n_2}$ has i.i.d. mean zero Gaussian entries $\epsilon_{ij} \sim N (0, \sigma_\epsilon^2)$.
	 There are three settings of $\sigma_\epsilon$, and they are chosen such that the signal-to-noise ratios (SNR$:= (\E\|\bAz\|_F^2/\E\|\bm \epsilon\|_F^2)^{1/2}$) are 1, 5 and 10.

	We consider three different missing mechanisms and generate observation indicator matrix $\bm T$ from $\bm \Pi = (\pi_{ij})_{i,j=1}^{n_1,n_2}$ that are specified as follows:
   
	Setting 1: This setting is a uniform missing setting $\pi_{ij} = 0.25$ for all $i,j= 1,\dots, 200$.

	Setting 2: In this setting, we relate the missingness with the value of the target matrix.
	 For entries that have high values, they are more likely to be observed. More specifically, we set
	\begin{equation*}
	\pi_{ij}=
	   \begin{cases}
		 1/16, & \mbox{if } A_{\star, ij}  \leq q_{0.25} \\
		 0.25, &  \mbox{if } q_{0.25}  < A_{\star, ij}\leq q_{0.75} \\
		 7/16, & \mbox{if } A_{\star, ij}> q_{0.75}\\
	   \end{cases}
	 \end{equation*}
   where $q_a$ is the $a$ quantile of $A_{\star,ij}$, $i,j = 1,\dots, 200$.
   
	Setting 3: This setting is the contrary of Setting \revise{2}. For entries that have high values, they are less likely to be observed.
	\begin{equation*}
	   \pi_{ij}=
		  \begin{cases}
		   7/16, & \mbox{if } A_{\star, ij}  \leq q_{0.25} \\
			0.25, &  \mbox{if } q_{0.25}  <A_{\star, ij}  \leq q_{0.75} \\
			1/16, & \mbox{if } A_{\star, ij} > q_{0.75}\\
		  \end{cases}
		\end{equation*}
	  where $q_a$ is the $a$ quantile of $A_{\star,ij}$, $i,j = 1,\dots, 200$.

	 \revise{We generate 200 simulated data sets separately for each of the above settings to compare different matrix completion methods, including
	 the proposed method (\proposed{}) and
	 five existing matrix completion methods: \citet{Mazumder-Hastie-Tibshirani10} (\unitrace), \citet{Cai-Zhou16} (\unimax), \citet{Fang-Liu-Toh18} (\unihy), \citet{Koltchinskii-Lounici-Tsybakov11} (\KLT) and  \citet{Negahban-Wainwright12} (\NW).}
	   For all methods mentioned above, we randomly separate $20\%$ of the observed entries in every simulated dataset and use it as the validation set to select tuning parameters.

	 In addition to the empirical root mean squared error (RMSE), we also include estimated rank and test error:
	 $$
	 \mbox{TE} :=\frac{\| (\bm J-\bm T) \circ (\widetilde{ \bm A} -\bAz)\|_F}{\sqrt{n_1n_2-N}},
	 $$
	 where $\widetilde{\bm A}$ is a generic estimator of $\bAz$; $\bm T$ is  the matrix of observed indicator and $N$ is the number of observed entries.
	 The test error measures the relative estimation error of the unobserved entries. Due to the space limitation, we only present the results for SNR = 5. Results for SNR = 1 and SNR = 10 can be found in \supp{Section \ref{appsec:sim} of the supplemental document}.  Table \ref*{tab:sim_SNR5} summarizes the average RMSE, average TE, and average estimated ranks for all three settings.
	  In all three settings, \unitrace{}, \unimax{} and \KLT{} do not provide competitive results as others. For Setting 1, \NW{} achieves the smallest RMSE and TE, but \proposed{}
	  performs closely to it. When SNR = 1 (shown in supplemental document), \proposed{} performs best --- the average RMSE of \proposed{} is 1.901 while the average RMSE of \NW{} is  2.012.
	  As for Settings 2 and 3,
	 \proposed{} outperforms other methods. Also, \NW{} performs significantly worse than \proposed{} in Setting 2. 
	  \unihy{} has average RMSE and TE that are close to \proposed{} in Setting 2 but does not perform well in Setting 3. 
	  As a result, we can see that \proposed{} is quite robust across different missing structures.

   \begin{table}[ht]
	 \centering
	 \small
	 \caption{Simulation results for three Settings when SNR=5. The average RMSE ($\overline{\mbox{RMSE}}$), average TE ($\overline{\mbox{TE}}$), and average estimated ranks (\=r) with standard errors (SE) in parentheses are provided for six methods (\proposed{}, \unitrace{}, \unimax{}, \unihy{}, \NW{} and \KLT{}) in comparison. For the columns related $\overline{\mbox{RMSE}}$ and $\overline{\mbox{TE}}$, we bold results with the first two smallest errors.}
   \begin{tabular}{c|ccc}
   \hline 
   & \multicolumn{3}{c}{Setting 1}\tabularnewline
   Method  & $\overline{\mbox{RMSE}}$ & $\overline{\mbox{TE}}$ & \=r\\
   \hline
   \proposed{} & \textbf{0.679(0.001)} & \textbf{0.700(0.001)} & 25.150(0.128)  \\
	 \unitrace & 0.699(0.001) & 0.721(0.001) & 45.005(0.161) \\ 
	 \unimax & 0.895(0.002) & 0.899(0.002) & 51.075(0.121) \\ 
	\unihy & 0.682(0.001) & 0.703(0.001) & 26.705(0.131)\\ 
	\NW & \textbf{0.668(0.001)}&\textbf{0.688(0.001)}&28.04(0.187)\\
	 \KLT & 1.913(0.003) & 1.976(0.003) & 8.720(0.060)  \\ 
   \hline

   & \multicolumn{3}{c}{Setting 2}  \tabularnewline
   Method  & $\overline{\mbox{RMSE}}$ & $\overline{\mbox{TE}}$ & \=r \\
   \hline
   \proposed{} & \textbf{0.624(0.001)} & \textbf{0.635(0.001)} & 24.980(0.136)  \\ 
	 \unitrace & 0.648(0.001) & 0.660(0.001) & 41.240(0.104)\\ 
	 \unimax & 0.922(0.002) & 0.945(0.002) & 47.170(0.156)\\ 
	 \unihy & \textbf{0.628(0.001)} & \textbf{0.640(0.001)} & 26.045(0.145) \\ 
	 \NW & 0.665(0.002) & 0.674(0.002) & 22.030(0.806)\\
	 \KLT & 1.980(0.006) & 1.880(0.004) & 1.355(0.141)  \\ 

   \hline
   
   & \multicolumn{3}{c}{Setting 3} \tabularnewline
   Method &$\overline{\mbox{RMSE}}$ & $\overline{\mbox{TE}}$  &\=r\\
   \hline
   \proposed{}  & \textbf{0.925(0.002)} & \textbf{1.002(0.002)} & 24.090(0.138) \\ 
	 \unitrace & 1.143(0.003) & 1.254(0.003) & 47.240(0.144) \\ 
	 \unimax & 1.222(0.003) & 1.324(0.003) & 50.590(0.151) \\ 
	 \unihy  & 1.026(0.002) & 1.118(0.003) & 32.440(0.131) \\ 
	 \NW & \textbf{0.964(0.002)} & \textbf{1.043(0.002)} & 18.350(0.319)\\
	 \KLT  & 3.174(0.006) & 3.477(0.006) & 9.575(0.093) \\ 
   \hline
   
   \end{tabular}
	 \label{tab:sim_SNR5}
   \end{table}

\section{Real Data Applications}
\label{sec:real}
We applied the above methods to two real datasets:

1. Coat Shopping Dataset,  which is available at \url{http://www.cs.cornell.edu/~schnabts/mnar/}. As described in \citet{Schnabel-Swaminathan-Singh16},  the dataset contains ratings from 290 Turkers on an inventory of 300 items. The self-selected ratings form the training set and the uniformly selected ratings form the test set. The training set consists of 6960 entries and test set consists of 4640 entries.

2. Yahoo!~Webscope Dataset,  which is available at \url{http://research.yahoo.com/Academic Relations}. It contains (incomplete) ratings from 15,400 users on 1000 songs. The dataset consists of two subsets, a training set and a test set. The training set records approximately 300,000 ratings given by the aforementioned 15,400 users. Each song has at least 10 ratings. The test set was constructed by surveying 5,400 out of these 15,400 users, such that each selected user rates exactly 10 additional songs.

For the second dataset, due to its large size, we use a non-convex algorithm of \citet{lee2010practical} to obtain \unimax{}.  Also, we modify this algorithm to incorporate another nuclear-norm regularization,  to obtain \proposed{} and \unihy{}. Detailed algorithm can be found in \supp{Section \ref{appsec:algo_nonconvex}  of the supplemental document.}  For both datasets, we separate half of the test data set as the validation set to select tuning parameters for all methods. And the remaining half test data set is used as the evaluation set. 

Here, we include the test root mean squared error
$$\mbox{TRMSE} := \frac{\|\bm {T}_e \circ(\widetilde{\bA}- \bAz)\|_F}{\sqrt{N_e}},$$
 where $\widetilde{\bm A}$ is a generic estimator of $\bAz$; $\bm {T}_e$ is the indicator matrix for the evaluation set and $N_e$ is the number of evaluation entries, and the test mean absolute error
 $$\mbox{TMAE} := \frac{\sum_{\bm{T}_{e, ij}=1}|\widetilde{\bA}_{ij}- \bm{A}_{\star, ij}|}{N_e},$$
 to measure the performance of all the methods. Rank estimation is also provided.

\begin{table}[t]
    \centering
    \caption{Test root mean squared errors (TRMSE), test mean absolute errors (TMAE) and estimated ranks (Rank) based on the evaluation set of Coat Shopping Dataset and Yahoo!~Webscope Dataset for \proposed{} and five existing methods proposed respectively in \citet{Mazumder-Hastie-Tibshirani10} (\unitrace), \citet{Cai-Zhou16} (\unimax), \citet{Fang-Liu-Toh18}(\unihy),  \citet{Negahban-Wainwright12} (\NW) and \citet{Koltchinskii-Lounici-Tsybakov11} (\KLT).
     For the columns related TRMSE and TMAE, we bold results with the first two smallest errors.}
 
  \begin{tabular}{c|ccc}
  \hline 
  & \multicolumn{3}{c}{Coat Shopping Dataset}\tabularnewline

  Method  & TRMSE & TMAE & Rank\\
  \hline
  \proposed{} & \textbf{0.9888} & \textbf{0.7627} & 26 \\
    \unitrace & 1.1401  & 0.8485  & 15 \\ 
    \unimax & 1.0354 & 0.8279 & 31 \\ 
   \unihy & \textbf{0.9980} & \textbf{0.7723} & 32\\ 
   \NW &  1.0553 & 0.7972&25\\
    \KLT & 2.0838 &  1.5733 & 2 \\ 
  \hline\hline
  & \multicolumn{3}{c}{Yahoo!~Webscope Dataset}  \tabularnewline
  Method  & TRMSE & TMAE & Rank\\
  \hline
  \proposed{} & \textbf{1.0111} &\textbf{0.7739} & 64 \\ 
    \unitrace & 1.2172 & 0.9230  & 31\\ 
    \unimax & 1.0339 & 0.8156 & 29\\ 
    \unihy & 1.0339 & 0.8156 & 29 \\ 
    \NW &  \textbf{1.0338}  &  \textbf{0.7954}  & 25\\
    \KLT &  3.811 &  1.6589 & 1  \\ 
  \hline

  \end{tabular}
    \label{tab:real}
  \end{table}

Table \ref{tab:real} shows the TRMSE, TMAE  and estimated ranks for the two datasets with all the methods mentioned above. For Coat Shopping Dataset, compared with the existing methods, the proposed method \proposed{} achieves best TRMSE and TMAE. The errors of \unihy{} are similar to that of \proposed{}, but the estimated rank is larger than that of \proposed{}. 
In other words, \proposed{} is significantly more efficient in capturing the signal.
For Yahoo!~Webscope Dataset, \proposed{} also has the smallest errors among all the methods. However, compared with \unimax{}, \unihy{} and \NW{} whose errors are relatively close to that of \proposed{}, \proposed{} has a higher estimated rank, though $64$ is a reasonably small rank for a matrix with size 1000 by 15400.
To confirm the fact that the higher errors of \unimax{}, \unihy{} and \NW{} are not due to
their smaller rank estimates,
we look into the test error sequences obtained by varying the tuning parameters, for each of these three methods. We find that the change of test errors (based on the evaluation set) aligns well with the validation errors (based on the validation set), and the chosen tuning parameters indeed correspond to the almost smallest test errors they can achieve.
This suggests that these three estimators are not able
to capture additional useful information and hence produce a smaller rank estimates.
But the proposed estimator is able to capitalize these additional signals to achieve reduction in errors.

\section*{Acknowledgements}
The authors thank the reviewers for their helpful comments and suggestions. The work of Raymond K. W. Wong is partially supported by the US National Science Foundation (DMS-1711952, DMS-1806063 and CCF-1934904). The work of Xiaojun Mao is partially supported by NSFC Grant No. 12001109 and 92046021, Shanghai Sailing Program 19YF1402800, and the Science and Technology Commission of Shanghai Municipality grant 20dz1200600.  The work of K. C. G. Chan is partially supported by the US National Science Foundation (DMS-1711952).
Portions of this research were conducted with high performance research computing resources provided by Texas A\&M University (\url{https://hprc.tamu.edu}).

\bibliographystyle{chicago}
\bibliography{MC_Balancing.bib}

\makeatletter\@input{xx1.tex}\makeatother
\end{document}


\def\spacingset#1{\renewcommand{\baselinestretch}%
{#1}\small\normalsize}

\linsps

\title{\bf Supplementary Material for ``Matrix Completion with Model-free Weighting"}

\author[1]{Jiayi Wang}
\author[1]{Raymond K. W. Wong}
\author[2]{Xiaojun Mao}
\author[3]{Kwun Chuen Gary Chan}
\affil[1]{Department of Statistics, Texas A\&M University}
\affil[2]{School of Data Science, Fudan University}
\affil[3]{Department of Biostatistics, University of Washington}
\date{}

\maketitle

\section{Proof of Lemma \ref{lem:BC}}
\label{appsec: BC}
\begin{proof}[Proof of Lemma \ref{lem:BC}]
  We first list two properties of max norm as follows.
   (i) As shown in \citet{Srebro-Shraibman05}, $\|\bB\|_*\le \sqrt{n_1n_2} \|\bB\|_{\max}$.
  (ii) 
  By an equivalent definition of max norm due to \citet{Lee-Shraibman-Spalek08} (also see equation (8) in \citet{Jalali-Srebro12}), we have $\|\bC\circ \bB\|\le \|\bC\|\|\bB\|_{\max}$.
   Together with the duality of nuclear norm, we can show that
  \begin{align}
    \label{eqn:optomax}
    |\langle \bC \circ \bB, \bB\rangle|&\le \|\bC\circ \bB\| \|\bB\|_*
    \leq \|\bC\| \|\bB\|_{\max}\|\bB\|_*\\
   & \le \sqrt{n_1 n_2}\|\bC\| \|\bB\|_{\max}^2. \nonumber
  \end{align}
\end{proof}

\section{Proofs of Theorems \ref{thm:berr}, \ref{thm:max} and \ref{thm:unilowerbound}}

  Let $\bm{e}_{i}(n)\in\mbR^n$ be the canonical basis vector, i.e., the $i$-th element of $\bm{e}_i(n)$ is $1$ and the remaining elements are $0$.
  We can define similar standard basis elements for $n_1$-by-$n_2$ matrices:
	 $\bJ_{ij}=\bm{e}_{i}(n_1)\bm{e}^\intercal_{j}(n_2)$,
   which will be used in the applications of matrix Bernstein
   inequality in our proofs. For any $\beta\ge 0$,
   define the class of matrices $\mathcal{B}_{\max}(\beta)$
   to be the max-norm ball with radius $\beta$, i.e.,
   \[
   \mathcal{B}_{\max}(\beta) = \{\bA\in\mbR^{n_1\times n_2}:\norm{\bA}_{\max}\le \beta\}.
   \]
   We also define
   \[
     \mathcal{F} = \{\bu\bv^T : \bu\in \{-1,+1\}^{n_1}, \bv\in \{-1,+1\}^{n_2}\},
   \]
   the set of rank-one sign matrices.
   Denote by $K_{G}\in(1.67,1.79)$ the Grothendieck's constant. From \citet{Srebro-Shraibman05},
   \begin{equation}\label{appeqn:SSconv}
   \text{conv}\mathcal{F} \subseteq\mathcal{B}_{\max}(1)\subseteq K_{G}\text{conv}\mathcal{F}.
   \end{equation}
	Moreover, the cardinality of $\mathcal{F}$ is $\abs{\mathcal{F}}=2^{n_1+n_2-1}$.

\begin{lem}\label{applem:SWA}
	Suppose Assumption \ref{ass:theta} hold.
	Let $\bW_{\diamond}=
	(w_{\diamond,i,j})\in\mathbb{R}^{n_1,n_2}$ where $W_{\diamond,i,j}=\pi_{i,j}^{-1}$.
	There exists a constant $\oldcnt{cnt:SWA}\ge0$ such that with probability at least $1-1/(n_1+n_2)$,
	\[
	\frac{1}{n_1n_2}\Norm{\bT\circ\bW_{\diamond}-\bJ}
	\le \oldcnt{cnt:SWA} \min\left\{\frac{\log^{1/2}(n_1+n_2)}{\sqrt{\pi_{L}(n_1\wedge n_2)n_1n_2}},\frac{\sqrt{n_1+n_2}}{\pi_{L}n_1n_2}\right\}.
	\]
\end{lem}

\begin{proof}[Proof of Lemma \ref{applem:SWA}]
  We use two different proof techniques to show the bounds.  Depending on the rate of $\pi_L$, one of these two bounds is faster. 

  First, we show the proof for deriving the second bound. 
  As $\{T_{ij}\}$ are Bernoulli random variables, each entry $T_{ij}W_{\diamond,i,j}-1$ of matrix $\bT\circ\bW_{\diamond}-\bJ$ is sub-Gaussian random \revise{variable}. Thus according to the definition of \revise{the} $\psi_2$ norm, we have
	\[
	\E\exp\{\log(2)\cdot(T_{ij}W_{\diamond,i,j}-1)^2/(\pi_{ij}^{-1}-1)^2\}\le2,
	\]
	which implies that $\norm{T_{ij}W_{\diamond,i,j}-1}_{\psi_2}\le\log^{-1/2}2\cdot(\pi_{ij}^{-1}-1)\le 2(\pi_{ij}^{-1}-1)$.

  By \revise{Theorem \ref{prop:4.4.5} in Section \ref{sec:useful_props}} 
  , taking $K=\max_{i,j}\norm{T_{ij}W_{\diamond,i,j}-1}_{\psi_2}\le2\pi_{L}^{-1}$ and $t=(n_1+ n_2)^{1/2}$ \revise{ in Theorem \ref{prop:4.4.5}}, there exists an absolute constant $\oldcnt{cnt:SWA}>0$ such that
	\[
	\Norm{\bT\circ\bW_{\diamond}-\bJ}\le \frac{\oldcnt{cnt:SWA}\sqrt{n_1+n_2}}{\pi_{L}},
	\]
	with probability at least $1-2\cdot\exp(-(n_1+ n_2))$.

  Next, we consider applying the Matrix Bernstein inequality to derive the first bound. 
  For $(n_1n_2)^{-1}\norm{\bT\circ\bW_{\diamond}-\bJ}=\norm{\sum_{i,j}(T_{ij}W_{\diamond,i,j}-1)\bJ_{ij}/(n_1n_2)}$, where $\bJ_{ij}$ has 1 for $(i,j)-$th, but 0 for all the remaining entries,  let \revise{$\bM_{i,j}=(T_{ij}W_{\diamond,i,j}-1)\bJ_{ij}$, $i = 1,\dots,n_1$, $j = 1,\dots,n_2$}, then \revise{$(n_1n_2)^{-1}\norm{\bT\circ\bW_{\diamond}-\bJ}=\norm{(n_1n_2)^{-1} \sum_{i,j}  \bM_{i,j}}$}.    We can easily verify that \revise{$\E(\bM_{i,j})=\bm{0}$ and $\norm{\bM_{i,j}}\le \max\{\pi_{L}^{-1}-1,1\}$} for each $i,j$ by Assumption \ref*{ass:theta}.
  
  Since $\E (T_{ij}W_{\diamond,i,j}-1)^2 = \pi_{ij}^{-1} - 1$, we can show that
\begin{align*}
  &\left\| \frac{1}{n_1n_2}  \sum_{i,j}  \E \left( \bM_{i,j}  \bM_{i,j}^\tp\right)\right\| =  \left\| \frac{1}{n_1n_2}  \sum_{i,j}  \E \left( \bM_{i,j}^{\tp}  \bM_{i,j}\right)\right\|\\
  \leq &\frac{1}{n_1n_2} \max\left\{\underset{1\le i \le n_1}{\max}\sum^{n_2}_{j=1}\Abs{1/\pi_{ij}-1} ,\underset{1\le j \le n_2}{\max}\sum^{n_1}_{i=1}\Abs{1/\pi_{ij}-1} \right\}\\
   \leq  &\frac{1}{n_1 \wedge n_2} |1/\pi_{L}-1|,
\end{align*}
where the first inequality comes from  
\revise{Corollary 2.3.2 in \citet{golub1996matrix}}.
	
	By \revise{Theorem \ref{prop:prop1} in Section \ref{sec:useful_props}} 
  , with probability at least $1-1/(n_1+n_2)$, we have
	\[
	\frac{1}{n_1n_2}\Norm{\bT\circ\bW_{\diamond}-\bJ}\le 2\max \left\{\sqrt{\frac{2\left|1/\pi_{L}-1\right|\log\left(n_1+n_2\right)}{\left(n_1 \wedge n_2\right)n_1n_2}},2\max\left\{\frac{1}{\pi_{L}}-1,1\right\}\frac{\log^{3/2}\left(n_1+n_2\right)}{n_1n_2}\right\}.
  \]
  
  Overall, the conclusion follows.
\end{proof}

  \begin{lem}\label{applem:TWfro}
    Suppose Assumption \ref{ass:theta} holds.
    With probability at least
    $1-\exp\{-2^{-1}(\log 2) \pi_L^{2} \sum_{i,j} \pi_{ij}^{-1}\}$,
    \[
      \| \bm{T}\circ \bm{W}_\diamond\|_F^2 \le 2 \sum_{i,j} \pi_{ij}^{-1}.
    \]
    In particular, the probability is lower bounded by
    $1-\exp\{-2^{-1}(\log 2) n_1 n_2 \pi_L^2 \pi_{U}^{-1}\}$.
  \end{lem}

  \begin{proof}[Proof of Lemma \ref{applem:TWfro}]
    Note that $\| \bm{T}\circ \bm{W}_\diamond\|_F^2=\sum_{i,j} T_{ij} \pi_{ij}^{-2}$.
    Let $\xi >0$.
    By Markov inequality,  for any $t\ge 0$,
    \begin{align*}
      \Pr\left( \| \bm{T}\circ \bm{W}_\diamond\|_F^2 \ge t \right) =
      \Pr\left\{ \exp(\xi \| \bm{T}\circ \bm{W}_\diamond\|_F^2 )\ge \exp(\xi t) \right\}
      &\le \exp(-\xi t)\E\exp\left(\xi \sum_{i,j} T_{ij} \pi_{ij}^{-2}\right)\\
      &= \exp(-\xi t) \prod_{i,j} \E\exp(\xi T_{ij} \pi_{ij}^{-2}).
    \end{align*}
    For each $(i,j)$, due to the inequality $1+x \le \exp(x)$ for $x\ge 0$,
    \[
      \E\exp(\xi T_{ij}\pi_{ij}^{-2}) =  1 + \{\exp(\xi \pi_{ij}^{-2})-1\}\pi_{ij}
      \le \exp[ \{\exp(\xi \pi_{ij}^{-2})-1\}\pi_{ij}].
  \]
  Combining with the above result and taking $t=2\sum_{i,j}\pi_{ij}^{-1}$,
  \begin{align*}
    \Pr\left( \| \bm{T}\circ \bm{W}_\diamond\|_F^2 \ge 2\sum_{i,j} \pi_{ij}^{-1}\right)
      &\le
      \exp\left[
        -2\xi \sum_{i,j}\pi_{ij}^{-1}
        + \sum_{i,j}\{\exp(\xi \pi_{ij}^{-2})-1\}\pi_{ij}
      \right]\\
      &=
      \exp\left[-\sum_{i,j} {\pi_{ij}}\left\{
          1+ 2\xi\pi_{ij}^{-2}
        - \exp(\xi \pi_{ij}^{-2})
  \right\}\right].
  \end{align*}
  Note the above inequality holds for any $\xi>0$.

  Next, we focus on the term $g(\xi \pi_{ij}^{-2})$ where
  $g(x) = 1+ 2 x - \exp(x)$ for $x\ge 0$.
  It is easy to show that $g$ attains its maximum at $x=\log 2$, and $g(\log 2) = 2\log 2 - 1>0$.
  Also, $g(x)$ is increasing for $0\le x \le \log 2$.

  Take $\xi=(\log 2) \pi_L^2$. Then $0 \le \xi \pi_{ij}^{-2}\le \log 2$,
  and hence $g(\xi \pi_{ij}^{-2}) >0$, for all $i,j$.
  The lower bound of $g(\xi \pi_{ij}^{-2})$ is crucial in determining the order of the probability bound.
  Since $g(x) \ge x/2$ for $0\le x\le \log 2$,
  \begin{align*}
    g(\xi \pi_{ij}^{-2}) = g( \pi_L^2 \pi_{ij}^{-2}\log 2)
    \ge  \frac{\log 2}{2}\pi_L^2 \pi_{ij}^{-2}, \quad \forall i,j
  \end{align*}
  We conclude that
  \begin{align*}
    \sum_{i,j} {\pi_{ij}}\left\{
          1+ 2\xi\pi_{ij}^{-2}
        - \exp(\xi \pi_{ij}^{-2})
      \right\}
      \ge\frac{\log 2}{2} \pi_L^2 \sum_{i,j} \pi_{ij}^{-1}
      \ge \frac{\log 2}{2} n_2 n_2 \pi_L^2 \pi_U^{-1},
  \end{align*}
  which leads to the desired result.
  \end{proof}

  With these two lemmas, we are posed to prove Theorem \ref{thm:berr}.
  \begin{proof}[Proof of Theorem \ref{thm:berr}]

  By Lemma \ref{applem:TWfro}, we can show that
  with probability at least $1-\exp\{-2^{-1}(\log 2) \pi_L^{2} \sum_{i,j} \pi_{ij}^{-1}\}$,
  $\|\bm{T}\circ \bm{W}_\diamond\|_F \le (2\sum_{i,j}\pi_{ij}^{-1})^{1/2}$
  and hence $\bm{W}_\diamond$ is feasible for the constrained optimization
  \eqref{eqn:minW}.

  Based on the definition  of the proposed estimator $\bhW$,  we have
  \begin{align*}
	S\left(\bhW,\bDe\right)
  &=\frac{1}{n_1n_2}\left|\Inner{\bDe}{\left(\bT\circ\bhW-\bJ\right)\circ\bDe}\right|\\
  &\le\frac{1}{\sqrt{n_1n_2}} \Norm{\bT\circ\bW_\diamond-\bJ}\Norm{\bDe}_{\max}^2\\
  &\le \frac{\beta'^2 }{\sqrt{n_1n_2}} \Norm{\bT\circ\bW_\diamond-\bJ}.
\end{align*}
  The desired result then follows from Lemma \ref{applem:SWA}.
  \end{proof}

  Our theoretical result of the final estimator $\bhA$ will be based on a key lemma (Lemma \ref{applem:dualcon}), which establishes the dual of max norm of random matrix $\bepsilon$ with general entry-wise scaling. Before we prove Lemma \ref{applem:dualcon}, we now show a comparison theorem between sub-Gaussian complexity and Gaussian complexity.
  This result (Lemma \ref{applem:BCF8}) extends Theorem 8 in \citet{Banerjee-Chen-Fazayeli14}
  to allow arbitrary entrywise scaling.

Define the Gaussian width and Gaussian complexity of the set $\mathcal{A}$ respectively as
\[
w(\mathcal{A})=\E_{\bG}\left[\sup_{\bA\in\mathcal{A}}\Inner{\bA}{\bG}\right]
\quad\mbox{and}\quad
\tilde{w}(\mathcal{A})=\E_{\bG}\left[\sup_{\bA\in\mathcal{A}}|\Inner{\bA}{\bG}|\right],
\]
where $\bG=(G_{ij})$ and each $\{G_{ij}\}$ are independent standard Gaussian random variables.
In our study, $\mathcal{A}$ is a max-norm ball, and so is symmetric. Therefore
Gaussian width and Gaussian complexity
are equivalent.

\begin{lem}[Extension of Theorem 8 in \citet{Banerjee-Chen-Fazayeli14}]\label{applem:BCF8}
  Suppose Assumption \ref{ass:error} holds.
  Let $\bB=(B_{ij})\in\mathbb{R}^{n_1\times n_2}$ be a fixed matrix such that $B_{ij}\ge 0$ for each $i,j$.
  Then
	\begin{equation*}
    \E \left[ \|\bB\circ \bepsilon\|_{\max}^* \right] \le \eta_0 \tau \E \left[
      \|\bB\circ\bm{G}\|_{\max}^*
    \right],
	\end{equation*}
  where $\|\cdot\|_{\max}^*$ is the dual norm of max norm, $\bG=(G_{ij})$ has independent standard Gaussian entries which are also independent of the random errors $\{\epsilon_{ij}\}$,
  and $\eta_0>0$ is an absolute constant.
\end{lem}

\begin{proof}[Proof of Lemma \ref{applem:BCF8}]
  Since the desired result obviously holds if $\bB=\bm 0$,
  we assume $\bB\neq \bm{0}$ in the rest of this proof.
   By definition, $\|\bC\|_{\max}^* = \sup_{\|\bm X\|_{\max}\le 1} \langle \bm X, \bm C\rangle$ 
  for any $\bC\in\mathbb{R}^{n_1\times n_2}$.
  Therefore our goal is to bound a scaled sub-Gaussian complexity via the corresponding scaled Gaussian complexity.
	We now extend the proof of Theorem 8 of \citet{Banerjee-Chen-Fazayeli14} to
  allow an additional entrywise scaling parameter $\bB$.
  We start with considering the sub-Gaussian process $Y_{\bm{X}} =
  \inner{\bm{X}}{\bB\circ \bepsilon}$ and the Gaussian process $Z_{\bm{X}} =
  \inner{\bm{X}}{\bB\circ \bG}$,
  both indexed by $\bm{X} \in \mathcal{B}_{\max}(1)$.
  For any $\bm{X}_1,\bm{X}_2\in\mathcal{B}_{\max}(1)$, by the general Hoeffding's inequality given in Theorem 2.6.3 of \citet{Vershynin18}, we have
	\begin{equation}
    \Pr\left(|Y_{\bm{X}_1} - Y_{\bm{X}_2}| \geq t \right) \leq 2 \cdot \exp\left( - \frac{\newcnt\ltxlabel{cnt:BCF8} t^2}{\tau^2\Norm{\bB\circ\left(\bm{X}_1 -\bm{X}_{2}\right)}_{F}^2}\right), \quad t>0,
	\end{equation}
  where $\oldcnt{cnt:BCF8} > 0$ is an absolute constant. One can show that $\E(Z_{\bm{X}_1}-Z_{\bm{X}_2})^2=\norm{\bB\circ(\bm{X}_1 - \bm{X}_2)}_{F}^2$. According to Theorem 2.1.5 of \citet{Talagrand06}, we can apply the generic chaining argument for upper bounds on the empirical processes $\sqrt{c} Y_{\bm X}/\tau$ and $Z_{\bm X}$. This yields
	\begin{equation}
	\E_{\bepsilon}\left[ \sup_{\bm{X}_1,\bm{X}_2\in\mathcal{B}_{\max}(1)} | Y_{\bm{X}_1} - Y_{\bm{X}_2} |\right] \leq \eta_1\tau \E_{\bG}\left[ \sup_{\bm{X}_1\in\mathcal{B}_{\max}(1)} Z_{\bm{X}_1} \right] = \eta_1\tau w(\mathcal{B}_{\max}(1)),
	\end{equation}
  where $\eta_1$ is an absolute constant.
  Further, we can see that if $\bm X\in\mathcal{B}_{\max}(1)$, then $-\bm X\in\mathcal{B}_{\max}(1)$.
  Then we have
  \begin{align*}
  \sup_{{\bm{X}_1},{\bm{X}_2}\in\mathcal{B}_{\max}(1)} | Y_{\bm{X}_1} - Y_{\bm{X}_2} |&=\sup_{{\bm{X}_1},{\bm{X}_2}\in\mathcal{B}_{\max}(1)} ( Y_{\bm{X}_1} - Y_{\bm{X}_2})=\sup_{{\bm{X}_1}\in\mathcal{B}_{\max}(1)}  Y_{\bm{X}_1} +\sup_{{\bm{X}_2}\in\mathcal{B}_{\max}(1)} ( - Y_{\bm{X}_2})\\
  =&\sup_{{\bm{X}_1}\in\mathcal{B}_{\max}(1)}  Y_{\bm{X}_1} +\sup_{-{\bm{X}_2}\in\mathcal{B}_{\max}(1)} ( \inner{-\bm{X}_2}{\bB\circ \bepsilon})=2\sup_{{\bm{X}_1}\in\mathcal{B}_{\max}(1)}  Y_{\bm{X}_1}.
  \end{align*}
  By taking the expectation on $\bepsilon$ on both side, we have
	\begin{equation}
	\E_{\bepsilon}\left[ \sup_{{\bm{X}_1},{\bm{X}_2}\in\mathcal{B}_{\max}(1)} | Y_{\bm{X}_1} - Y_{\bm{X}_2} | \right] = 2 \E_{\bepsilon}\left[ \sup_{\bm{X}_1\in\mathcal{B}_{\max}(1)} Y_{\bm{X}_1} \right].
	\end{equation}
	As a result, with $\eta_0 = \eta_1/2$, we have
	\begin{equation}
	\E_{\bepsilon}\left[\sup_{\bm{X}\in\mathcal{B}_{\max}(1)} \Inner{\bB\circ \bepsilon}{\bm{X}}\right] = \E_{\bepsilon}\left[ \sup_{\bm{X}\in\mathcal{B}_{\max}(1)} Y_{\bm{X}} \right] \leq \eta_0\tau w(\mathcal{B}_{\max}(1)).
	\end{equation}
	That completes the proof.
\end{proof}

\begin{lem}\label{applem:dualcon}
  Suppose Assumption \ref{ass:error} holds.
  Let $\bB=(B_{ij})\in\mathbb{R}^{n_1\times n_2}$ be a fixed matrix such that $B_{ij}\ge 0$ for each $i,j$. There exists an absolute constant $\newcnt\ltxlabel{cnt:dualnormB}>0$ such that, with probability at least $1-2\exp\{-(n_1+n_2)\}$,
\begin{equation*}
  \|\bB \circ \bepsilon\|_{\max}^* \le \oldcnt{cnt:dualnormB} \tau \|\bB\|_F\sqrt{n_1+n_2}.
\end{equation*}
\end{lem}

\begin{proof}[Proof of Lemma \ref{applem:dualcon}]
	Define the set
	\[
	\widetilde{\mathcal{B}}_{\max}(\beta) = \{\bB\circ \bm{X}: \bm{X}\in \mathcal{B}_{\max}(\beta)\} \subset \mathbb{R}^{n_1\times n_2}.
	\]
	Note that we have
\[
\E_{\bG} \left[ \sup_{\bm{X}\in\mathcal{B}_{\max}(1)} \Inner{\bB \circ \bG}{\bm{X}}\right] = \E_{\bG} \left[\sup_{\bm{X}\in\mathcal{B}_{\max}(1)} \Inner{\bG}{\bB \circ \bm{X}}\right] = w(\widetilde{\mathcal{B}}_{\max}(1)).
\]
Write $\widetilde{\mathcal{F}} = \{\bB \circ \bm{X}: \bm{X}\in\mathcal{F}\}$. By the the relationship \eqref{appeqn:SSconv}, we have
\[
\widetilde{\mathcal{F}}
\subseteq \widetilde{\mathcal{B}}_{\max}(1)
\subseteq \{\bB\circ \bm{X}: \bm{X}\in K_G\, \mathrm{conv}\,(\mathcal{F})\}
= K_G\{\bB \circ \bm{X}: \bm{X}\in \mathrm{conv}(\mathcal{F})\}
= K_G\,\mathrm{conv}(\widetilde{\mathcal{F}}).
\]
Due to the properties of Gaussian width  \citep[see, e.g., Appendix A.1 of][]{Banerjee-Chen-Fazayeli14}, we have
\[
w(\widetilde{\mathcal{B}}_{\max}(1)) \le w(K_G\,\mathrm{conv}(\widetilde{\mathcal{F}})) =K_G w(\mathrm{conv}(\widetilde{\mathcal{F}}))= K_G w(\widetilde{\mathcal{F}}).
\]

As for any $\bm{X}\in \mathcal{F}$, we have
\[
\|\bB\circ \bm{X}\|_F=\|\bB\|_F,
\]
and so $\langle \bm{G}, \bB\circ\bm X\rangle\sim \mcN(0,\|\bB\|_F^2)$.
Recall that the $|\mathcal{F}|= 2^{n_1+n_2-1}$.
By Proposition 3.1(ii) of \citet{Koltchinskii11}, we have
\[
  w(\widetilde{\mathcal{F}})\le \newcnt\ltxlabel{cnt:gaussianwidthfinite} \|\bB\|_F\sqrt{n_1+n_2}.
\]
where $\oldcnt{cnt:gaussianwidthfinite}$ is a absolute constant.
By Lemma \ref{applem:BCF8}, we conclude that
\begin{align}
\E_{\bepsilon} \left[ \sup_{\bm{X}\in\mathcal{B}_{\max}(1)} \Inner{\bB \circ \bepsilon}{\bm{X}}\right]
&\le\eta_0\tau\E_{\bG} \left[ \sup_{\bm{X}\in\mathcal{B}_{\max}(1)} \Inner{\bB \circ \bG}{\bm{X}}\right]\nonumber\\
&=\eta_0 \tau w(\widetilde{\mathcal{B}}_{\max}(1))
\le K_G\eta_0\tau w(\widetilde{\mathcal{F}})\nonumber \\
&\le \oldcnt{cnt:gaussianwidthfinite} K_G\eta_0 \tau \|\bB\|_F \sqrt{n_1+n_2}.\label{eqn:maxexpectbound}
\end{align}

Let $\varphi(\bm{Z})=\sup_{\|\bm{X}\|_{\max}\le 1}\langle \bB\circ \bm{Z}, \bm X\rangle$
for any $\bm{Z}\in \mathbb{R}^{n_1\times n_2}$.
We aim to provide the concentration of $\varphi(\bepsilon)$
to its expectation.
For notational simplicity, we will focus on the setting with $B_{ij}>0$ for all $i,j$;
otherwise, one can reduce the support of $\varphi$ to those entries corresponding to non-zero $B_{ij}$.
Due to the possibly unbounded support of $\bepsilon$,
we adopt an extension of McDiarmid's inequality \citet{Kontorovich14}
with unbounded diameter.
For any $\bm{Z}_1=(Z_{1,ij}), \bm{Z}_2=(Z_{2,ij})\in \mathbb{R}^{n_1\times n_2}$,
\begin{align*}
  \left|  \varphi(\bm{Z}_1)- \varphi(\bm{Z}_2) \right|
&\le \sup_{\|\bm X\|_{\max}\le 1}|\langle \bB\circ \bm Z_1, \bm X\rangle -\langle \bB\circ \bm Z_2, \bm X\rangle|\\
&\le \sup_{\|\bm X\|_{\max}\le 1} \sum_{i,j} B_{ij} |X_{ij}| |Z_{1,ij} - Z_{2,ij}|\\
&\le \sup_{\bm X\in K_G \mathrm{conv}(\mathcal{F})} \sum_{i,j} B_{ij} |X_{ij}| |Z_{1,ij} - Z_{2,ij}|\\
&\le q(\bm{Z}_1, \bm{Z}_2),
\end{align*}
where $q(\bm{Z}_1, \bm{Z}_2)=: \sum_{i,j} q_{ij}(Z_{1,ij}, Z_{2,ij})=: \sum_{i,j} K_G B_{ij} |Z_{1,ij} - Z_{2,ij}|$ is a metric.
Therefore $\varphi$ is 1-Lipschitz with respect to the metric $q$.
Let $\varepsilon_{ij}'$ be an independent copy of $\varepsilon_{ij}$,
and $\gamma_{ij}$ be an independent Rademacher random variable.
We can show that the subgaussian norm of $\gamma_{ij}q_{ij}(\varepsilon_{ij}, \varepsilon_{ij}')$
is bounded by $\newcnt\ltxlabel{cnt:dualcon}\tau b_{ij}$ for some absolute constant $\oldcnt{cnt:dualcon}>0$.
By Theorem 1 of \citet{Kontorovich14}, we conclude that
\[
  \pr(|\varphi(\bepsilon) - \E\varphi(\bepsilon)| > t)\le 2 \exp\left( -\frac{ t^2}{2\oldcnt{cnt:dualcon}\tau^2\|\bB\|_F^2}\right), \quad t\ge 0.
\]
Combining with \eqref{eqn:maxexpectbound}, we achieve the desired result.

\end{proof}

	\begin{lem}\label{applem:TWe}
	Suppose Assumptions \ref{ass:theta} and \ref{ass:error} hold.
  There exists an absolute constant $\oldcnt{cnt:dualnormB}>0$ such that with probability at least $1-2\exp\{-(n_1+n_2)\}$,
	\[
	\Norm{\bT\circ\bhW\circ\bepsilon}_{\max}^{\ast}
  \le \oldcnt{cnt:dualnormB} \tau\kappa \sqrt{n_1+n_2}.
	\]
	\end{lem}

	\begin{proof}[Proof of Lemma \ref{applem:TWe}]

    Notice that $\bepsilon$ is independent of $\bT$,
    and $\bhW$ is a function of $\bT$.
    By Lemma \ref{applem:dualcon},
    conditioned on $\bm{T}$, we have
    \begin{equation}
      \Norm{\bT\circ\bhW\circ\bepsilon}_{\max}^{\ast}
	  \le \oldcnt{cnt:dualnormB}\tau
    \|\bT\circ \bhW\|_F \sqrt{n_1 + n_2},
\label{appeqn:dualnormineq}
    \end{equation}
    with conditional probability at least $1-2\exp\{-(n_1+n_2)\}$.
     Since the probability bound does not depend on $\bT$,
     \eqref{appeqn:dualnormineq} holds with the same probability bound unconditionally.
  By construction, $\|\bT\circ \bhW\|_F\le \kappa$, we have the desired result.
	\end{proof}

\begin{proof}[Proof of Theorem \ref{thm:max}]

  It follows from the definition of $\bhA$ that for $\bAz\in\mbR^{n_1\times n_2}$ with $\|\bAz\|_{\max} \leq \beta$,
	\begin{equation}\label{appeqn:bydef}
	\frac{1}{n_1n_2}\Norm{\bT\circ\bhW^{\circ 1/2}\circ\left(\bhA-\bY\right)}_{F}^{2}\le\frac{1}{n_1n_2}\Norm{\bT\circ\bhW^{\circ 1/2}\circ\left(\bAz-\bY\right)}_{F}^{2} + \mu(\|\bAz\|_*  - \|\bhA\|_*).
	\end{equation}
	Since we can rewrite the first term in the left hand side of (\ref{appeqn:bydef}) as
	\[
	\frac{1}{n_1n_2}\Norm{\bT\circ\bhW^{\circ 1/2}\circ\left(\bhA-\bY\right)}_{F}^{2}=\frac{1}{n_1n_2}\Norm{\bT\circ\bhW^{\circ 1/2}\circ\left(\bhA-\bAz+\bAz-\bY\right)}_{F}^{2} ,
	\]
	the inequality (\ref{appeqn:bydef}) leads to
	\begin{align*}
	\frac{1}{n_1n_2}\Norm{\bT\circ\bhW^{\circ 1/2}\circ\left(\bhA-\bAz\right)}_{F}^{2}\le&\frac{2}{n_1n_2}\Inner{\bT\circ\bhW^{\circ 1/2}\circ\left(\bhA-\bAz\right)}{\bT\circ\bhW^{\circ 1/2}\circ\left(\bY-\bAz\right)} +  \mu(\|\bAz\|_*  - \|\bhA\|_*)\\
	=&\frac{2}{n_1n_2}\Inner{\bhA-\bAz}{\bT\circ\bhW\circ\bepsilon}+ \mu(\|\bAz\|_*  - \|\bhA\|_*).
  \end{align*}
  Therefore, due to  Theorem \ref{thm:berr}, Lemma \ref{applem:TWe} and condition of $\mu$, with the property that $\|\bAz\|_* \leq \sqrt{n_1n_2} \|\bAz\|_{\max}$,  we have
  \begin{align}\label{appeqn:final}
    \frac{1}{n_1n_2}\Norm{\bhA-\bAz}_{F}^{2}
    &\le\frac{1}{n_1n_2}\Inner{\bhA-\bAz}{\left(\bT\circ\bhW-\bJ\right)\circ\left(\bAz-\bhA\right)}+ \frac{1}{n_1n_2} \|\bm T \circ \ep{\bhW}{1/2}\circ (\bhA - \bAz)\|_F^2   \nonumber\\
    &\le S(\bhW, \bhA-\bAz) + \left|\frac{2}{n_1n_2} \langle \bhA - \bAz, \bT \circ \bhW \circ \bepsilon \rangle \right| + \mu (\|\bAz\|_* - \|\bhA\|_*)\nonumber\\
    &\le S(\bhW, \bhA-\bAz) + \frac{2}{n_1n_2}\Norm{\bhA-\bAz}_{\max}\Norm{\bT\circ\bhW\circ\bepsilon}_{\max}^{\ast} + \mu \|\bAz\|_* \nonumber\\
    & \le \oldcnt{cnt:SWA} (\beta^2 + \beta)  \min\left\{\frac{\log^{1/2}(n_1+n_2)}{\sqrt{\pi_{L}(n_1\wedge n_2)}},\frac{\sqrt{n_1+n_2}}{\pi_{L}\sqrt{n_1n_2}}\right\} + \frac{4\oldcnt{cnt:dualnormB}\beta\tau\kappa\sqrt{n_1+n_2}}{n_1n_2}\\
    & \le \oldcnt{cnt:SWA} (\beta^2)  \min\left\{\frac{\log^{1/2}(n_1+n_2)}{\sqrt{\pi_{L}(n_1\wedge n_2)}},\frac{\sqrt{n_1+n_2}}{\pi_{L}\sqrt{n_1n_2}}\right\} + \frac{4\oldcnt{cnt:dualnormB}\beta\tau\kappa\sqrt{n_1+n_2}}{n_1n_2}.
    \end{align}
with probability at least $1- \exp\{-2^{-1}(\log 2) \pi_L^{2} \sum_{i,j} \pi_{ij}^{-1}\} - 2\exp\{-(n_1+n_2)\} - 1/(n_1 + n_2)$.
\end{proof}

\begin{proof}[Proof of Theorem \ref{thm:unilowerbound}]
  Without loss of generality, we assume that $n_1\ge n_2$. 
  For some constant $0\le\gamma\le1$ such that 
  \revise{\(B =  \sigma^{-2}(\sigma \wedge \beta)^2  / (\gamma^2)\) is an integer and \(B \leq n_2\)}, define
  \revise{\[\mathcal{C}_1=\left\{\tilde{\bA}=\left(A_{ij}\right)\in\mathbb{R}^{n_1\times B}:A_{ij}\in\left\{0, \gamma\beta\right\},\forall 1\le i\le n_1,1\le j \le B\right\},\]}
	and consider the associated set of block matrices
	\[
	\mathcal{A}\left(\mathcal{C}_1\right)=\left\{\bA=\left(\widetilde{\bA}|\dots|\widetilde{\bA}|\bm{0}\right)\in\mathbb{R}^{n_1\times n_2}:\widetilde{\bA}\in\mathcal{C}_1\right\},
	\]
	where $\bm{0}$ denotes the $n_1\times (n_2-B\lfloor n_2/B\rfloor)$ zero matrix.

  It is easy to see that for any $\bA\in \mathcal{A}(\mathcal{C}_1)$, we have that $\norm{\bA}_{\max}\le\sqrt{B}\norm{\bA}_{\infty}\le\beta$. 
	Due to Lemma 2.9 in \citet{Tsybakov09}, there exists a subset $\mathcal{\bA}^{0}\subset\mathcal{A}(\mathcal{C}_1)$ containing the zero $n_1\times n_2$ matrix $\bm{0}$ where $\text{Card}(\mathcal{\bA}^{0})\ge 2^{Bn_1/8}+1$ and for any two distinct elements $\bA_{1}$ and $\bA_{2}$ of $\mathcal{A}^{0}$,

  \revise{
\begin{align}
  \label{eqn:distance}
  \Norm{\bA_{1}-\bA_{2}}_{F}^{2}\ge\frac{n_1B}{8}\left\{\gamma^2 \beta^2 \left\lfloor \frac{n_2}{B}\right\rfloor\right\}\ge \frac{n_1n_2\gamma^2 \beta^2}{16}.
\end{align}
  }

	For any $\bA\in\mathcal{\bA}^{0}$, \revise{from the noisy observed model in section 2.2, the probability distribution  $\mathbb{P}_{\bA}=\Pi_{i,j} [(2\pi\sigma^2)^{-1/2}\exp\{-(Y_{ij}-A_{ij})^2/(2\sigma^2)\}]^{T_{ij}}$. Take $\mathbb{P}_{\bm{0}}=\Pi_{i,j}[(2\pi\sigma^2)^{-1/2}\exp\{-Y_{ij}^2/(2\sigma^2)\}]^{T_{ij}}$. 
  Thus }the Kullback-Leibler divergence $K(\mathbb{P}_{\bm{0}},\mathbb{P}_{\bA})$\revise{$=\E_{\mathbb{P}_{\bm{0}}}(\log(\mathbb{P}_{\bm{0}}/\mathbb{P}_{\bm{A}}))$} between $\mathbb{P}_{\bm{0}}$ and $\mathbb{P}_{\bA}$ satisfies
  \revise{
    \begin{equation*}
      K\left(\mathbb{P}_{\bm{0}},\mathbb{P}_{\bA}\right)=\E_{\mathbb{P}_{\bm{0}}}\left(\sum_{ij}T_{ij}\frac{A_{ij}^2-2A_{ij}Y_{ij}}{2\sigma^2}\right)=\frac{\Norm{\bPi^{\circ 1/2}\circ\bA}_{F}^2}{2\sigma^2}
	\le \frac{\gamma^2 \beta^2 \sum_{i=1}^{n_1}\sum_{j=1}^{n_2}\pi_{ij}}{2\sigma^2} \le C_5\frac{\gamma^2 \beta^2 n_1 n_2 \pi_L}{2\sigma^2},
    \end{equation*}
  }
	for some positive constant $C_5$. \revise{ The last inequality is due to the condition that \(n_1n_2\pi_L \asymp\sum_{i=1}^{n_1}\sum_{j=1}^{n_2}\pi_{ij}.\)}

  From above we deduce the condition
	\begin{equation}
	\frac{1}{\text{Card}(\mathcal{\bA}^{0})-1}\sum_{\bA\in\mathcal{\bA}^{0}}K\left(\mathbb{P}_{\bm{0}},\mathbb{P}_{\bA}\right)\le \lambda \log\left(\text{Card}(\mathcal{\bA}^{0})-1\right),
  \end{equation}

 \revise{ The above condition is valid when we take  
  \[\gamma^2 = C_6 \left( \frac{ (\sigma \wedge \beta)^2}{\beta^2 n_2 \pi_L} \right)^{1/2}\]
for some constant \(C_6\) that  depends on \(\lambda\). Also, one can verify that  under the conditions \(\pi_L^{-1} = \bigO(\beta^2 (n_1\wedge n_2)/(\sigma\wedge \beta)^2)\) and \(\pi_L^{1/2} = \bigO((n_1\wedge n_2)^{1/2}\sigma^2 /[\beta (\sigma \wedge \beta))]\), \(\gamma \leq 1\) and \(B \leq n_2\). Then we subsitute \(\gamma^2\) in the bound of \ref{eqn:distance} and we achieve the final bound as the one showed in the Theorem. }

Together with the similar argument when $n_2\ge n_1$, the result now follows by application of Theorem 2.5 in \citet{Tsybakov09}. This completes the proof.
\end{proof}

\begin{lem}
  \label{lem:twe}
  Suppose Assumption \ref{ass:error} hold. For a fixed matrix $\bm B = (B_{ij})\in\mathbb{R}^{n_1\times n_2}$ where $B_{ij} \ge 0$, there exists an absolute  constant $\newcnt\ltxlabel{cnt:op}>0$ such that, with probability at least $1-2\exp(-(n_1 + n_2))$, 
  $$\|\bm B \circ \bepsilon\| \leq \oldcnt{cnt:op} \|\bm B\|_{\infty} \tau (\sqrt{n_1} + \sqrt{n_2}).$$
\end{lem}

\begin{proof}
  By the definition of $\|\cdot\|_{\psi_2}$,
  \begin{align*}
   \max_{i,j} \|B_{ij} \epsilon_{ij}\|_{\psi_2} \leq \|\bm B\|_\infty \tau.
  \end{align*}
Apply \revise{Theorem \ref{prop:4.4.5} in Section \ref{sec:useful_props}} 
, and take $t = \sqrt{n_1} + \sqrt{n_2}$ in \revise{Theorem \ref{prop:4.4.5}}. 
Then conclusion follows. 

\end{proof}

\section{Non-asymptotic Error Bound under Low-rank Settings and Asymptotically Homogeneous Missingness}
\label{sec:lr}
\begin{thm}
  \label{thm:uni_lr} 
  Suppose Assumption \ref{ass:error} hold and \(\pi_L {\asymp} \pi_U {\asymp} \pi\). Assume $\|\bAz\|_{\max}{\leq} \beta$ and \(\bAz\) has rank $R$. If \(\kappa' = \kappa - \|\bT\|_F\) is bounded and
$\mu {\asymp }\sqrt{{\{\tau^2\pi\log\left(n_1{+}n_2\right)\}}{\{\left(n_1 {\wedge} n_2\right)n_1n_2}\}^{-1}}$,
then there exists a constant $\newcnt\ltxlabel{cnt:hylr}>0$, such that with probability at least
$1{ - }{3}{(n_1{+}n_2)^{-1}}$,
\[ d^{2}(\bhA,\bAz) \le \frac{\oldcnt{cnt:hylr}R (\tau^2{\vee} \|A_\star\|_\infty^2 )\log(n_1 + n_2)}{[\pi (n_1{\wedge} n_2)]^{-1}}.\]
\end{thm}

\begin{proof}[Proof outline]
From the basic inequality, we have 
\begin{align*}
  \frac{1}{n_1n_2}\|\bT \circ \bhW^{1/2} \circ (\bhA-\bAz)\|_F^2 \leq \frac{2}{n_1n_2}\langle \bhA - \bAz, \bT \circ \bhW \circ \bm \epsilon \rangle + \mu\|\bAz\|_* - \mu \|\bhA\|_*.
\end{align*} 
Note that weights are restricted to be greater than 1. We then have 
\begin{align*}
  \frac{1}{n_1n_2}\|\bT \circ (\bhA-\bAz)\|_F^2 \leq  \frac{1}{n_1n_2}\|\bT \circ \bhW^{1/2} \circ (\bhA-\bAz)\|_F^2 \leq \frac{2}{n_1n_2}\|\bhA - \bAz\|_* \|\bT \circ \bhW \circ \bm \epsilon\|+ \mu(\|\bAz\|_* -  \|\bhA\|_*).
\end{align*}
Due to the constraint of \(\kappa'\), \(\|\bhW\|_\infty\) is bounded, we can use Lemma \ref{lem:twe} to derive the bound of \(\|\bT \circ \bhW \circ \bm \epsilon\|\). The remaining argument is rather standard and the same as the proof for No-weighted estimators with nuclear norm regularization \citep[][]{Klopp14}.  
\end{proof}

\begin{lem}
  \label{lem:twe}
  Suppose assumptions in Corollary \ref{thm:uni_lr} hold. Then there exists a  constant $\newcnt\ltxlabel{cnt:twefinal}>0$ such that, with probability at least $1-(n_1 + n_2)$, 
  $$\frac{1}{n_1n_2}\norm{\bT\circ \bhW \circ \bepsilon} \leq \oldcnt{cnt:twefinal} 
    \sqrt{\frac{\tau^2 \pi \log\left(n_1+n_2\right)}{\left(n_1 \wedge n_2\right)n_1n_2}}$$ 
\end{lem}

\begin{proof}
  The proof is very similar as the proof in Lemma \ref{applem:SWA}.

  We consider applying the Matrix Berinstein inequality for random matrices with bounded sub-exponential norm. 

  Due to the constraint of \(\kappa'\), there exists a constant $\newcnt\ltxlabel{w_upper}$
such that \(\|\hat{ \bm W}\|_\infty \leq \oldcnt{w_upper}\).

  For $(n_1n_2)^{-1}\norm{\bT\circ \bhW \circ \bepsilon}=\norm{\sum_{i,j}(T_{ij}\hat{W}_{i,j}\epsilon_{ij})\bJ_{ij}/(n_1n_2)}$, where $\bJ_{ij}$ has 1 for $(i,j)-$th, but 0 for all the remaining entries,  let $\bM_{i,j}=(T_{ij}\hat{W}_{i,j}\epsilon_{ij})\bJ_{ij}$, then $(n_1n_2)^{-1}\norm{\bT\circ \bhW \circ \bepsilon}=\norm{(n_1n_2)^{-1} \sum_{i,j}  \bM_{i,j}}$.    We can easily verify that $\E(\bM_{i,j})=\bm{0}$. Note that \(\epsilon_{ij}\) are sub-Gaussian random variables and therefore sub-exponential random variables. Then $\max_{i,j}\|\|\bm M_{i,j}\|\|_{\psi_1}\leq \max_{i,j}\|T_{ij}\hat{W}_{i,j}\epsilon_{ij}\|_{\psi_1} \leq \newcnt\ltxlabel{sub-exponential}\tau$, where $\|\cdot\|_{\psi_1}$ is the sub-exponential norm of a random variable and \(\oldcnt{sub-exponential}\) is some constant depending on the \(\oldcnt{w_upper}\). 
  
  Since $\E (T_{ij}\hat{W}_{i,j}\epsilon_{ij})^2 \leq c \oldcnt{w_upper}^2 \pi_{ij} \tau^2 $ for some absolute constant $c$, we can show that
\begin{align*}
  &\left\| \frac{1}{n_1n_2}  \sum_{i,j}  \E \left( \bM_{i,j}  \bM_{i,j}^\tp\right)\right\| =  \left\| \frac{1}{n_1n_2}  \sum_{i,j}  \E \left( \bM_{i,j}^{\tp}  \bM_{i,j}\right)\right\|\\
  \leq &\frac{1}{n_1n_2} \max\left\{\underset{1\le i \le n_1}{\max}\sum^{n_2}_{j=1} c_2 \pi_{ij} \tau^2 ,\underset{1\le j \le n_2}{\max}\sum^{n_1}_{i=1}c_2 \pi_{ij} \tau^2 \right\}\\
   \leq  &\frac{c_3\tau^2 }{n_1 \wedge n_2} \pi,
\end{align*}
for some constant \(c_3\).

	By Proposition 11 in \citet{Klopp14}
  , there exsits a constant $\oldcnt{cnt:twefinal} $, such that  with probability at least $1-1/(n_1+n_2)$,
	\[
	\frac{1}{n_1n_2}\Norm{\bT\circ \bhW \circ \bepsilon}\le \oldcnt{cnt:twefinal}  \max \left\{\sqrt{\frac{\tau^2 \pi \log\left(n_1+n_2\right)}{\left(n_1 \wedge n_2\right)n_1n_2}} ,   \tau \log (1/\sqrt{\pi})\frac{\log^{3/2}(n_1 + n_2)}{n_1 n_2}\right\}.
  \]
  
  Overall, the conclusion follows.
\end{proof}

\section{Useful Results}
\label{sec:useful_props}
\begin{thm}[Theorem 4.4.5 of \citet{Vershynin18}]
  \label{prop:4.4.5}
  Let \(\bm A\) be an \(n_1\times n_2\) random matrix whose entries \(A_{ij}\) are independent, mean zero, sub-gaussian random variables. Then, for any \(t >0\) we have
  \[\|\bm A\| \leq C K(\sqrt{n_1} + \sqrt{n_2} + t) \]
  with probability at least \(1-2\exp(-t^2)\). Here \(K = \max_{ij}\|A_{ij}\|_{\psi_2}\) and \(C\) is an absolute constant.
\end{thm}
\begin{proof}
  The proof can be found on Page 91 in \citet{Vershynin18}.
\end{proof}

\begin{thm}[Proposition 1 of \citet{Koltchinskii-Lounici-Tsybakov11}]
  \label{prop:prop1}
  Let \(\bm Z_1,\dots,\bm Z_N\) be independent random matrices with dimensions \(n_1\times n_2\) that satisfy \(\E \bm Z_i = 0\) and \(\| \bm Z_i\| \leq U\) almost surely for some constant \(U\) and all \(i = 1,\dots,n\). Define
  \[\sigma_Z = \max \left\{ \left\| \frac{1}{N}\sum_{i=1}^N \E(\bm Z_i \bm Z_i^\tp) \right\|^{1/2}, \left\| \frac{1}{N}\sum_{i=1}^N \E(\bm Z_i^\tp \bm Z_i) \right\|^{1/2}\right\}.\]
  Then, for all \(t >0\), with probability at least \(1-\exp(-t)\) we have 
  \[\left\|\frac{\bm Z_1 + \dots + \bm Z_N}{N}\right\| \leq 2 \max\left\{ \sigma_Z\sqrt{\frac{t + \log (n_1 + n_2)}{N}}, U \frac{t + \log (n_1 + n_2) }{N} \right\},\]
\end{thm}
  \begin{proof}
    The proof can be found on Page 2325 in \citet{Koltchinskii-Lounici-Tsybakov11}.
  \end{proof}

\section{Algorithm}
\subsection{Convex Algorithm for Solving \eqref{eqn:objA}}
\label{appsec:algo_convex}
Follow \citet{Fang-Liu-Toh18} and  \citet{Cai-Zhou16}, we consider an equivalent form objective function in \eqref{eqn:objA}  below.
\begin{align*}
  &\min_{\bm X,\bm Z} \frac{1}{n_1n_2} \|{\bT\circ \ep{\bhW}{1/2} \circ\left(\bY-\bm Z_{12}\right)}\|_{F}^2 + \mu \langle \bm I , \bm X \rangle, \\
  &\mathrm{Subject\  to\ } \revise{\bm X   \succcurlyeq 0 },\  \bm X = \bm Z,\  \bm Z \in \mathcal{P}_\beta
\end{align*}
where $\bm Z, \bm X \in\mathbb{R}^{({n_1+ n_2})\times (n_1+n_2)}$, $\mathcal{S}$ is the class of all symmetric matrices in  $\mathbb{R}^{({n_1+ n_2})\times (n_1+n_2)}$,  $\mathcal{P}_\beta := \{ \bm C \in \mathcal{S}: \mathrm{diag}(\bm C) \ge 0, \|\bm C\|_\infty \leq \beta
  \}$, $\bm I$ is an identity matrix and 
\[
\bm Z=
\left[
\begin{array}{cc}
\bm Z_{11} & \bm Z_{12} \\
\bm Z_{12}^{\tp} & \bm Z_{22}
\end{array}
\right], \bm Z_{11} \in \mathbb{R}^{n_1\times n_1}, \bm Z_{22} \in \mathbb{R}^{n_2\times n_2}
\]

\revise{The derivation of above representation mainly comes from two facts: 1. The nuclear norm of \(\bm Z_{12}\) is the the smallest possible sum of  elements on the diagonal of \(\bm Z\) given \(\bm Z \succcurlyeq 0\) \citep{fazel2001rank}; 2. The max norm of matrix \(\bm Z_{12}\) is the smallest possible maximum element on the diagonal of \(\bm Z\) given \(\bm Z \succcurlyeq 0\) \citep{srebro2005maximum}.} 

The augmented Lagrangian function can be written as
\begin{align*}
  \label{eqn: eq_admm}
  &\mathcal{L}(\bm X, \bm Z, \bm V) =  \frac{1}{n_1n_2} \|{\bT\circ \ep{\bhW}{1/2} \circ\left(\bY-\bm Z_{12}\right)}\|_{F}^2 + \mu \langle \bm I , \bm X \rangle +  \langle \bm V, \bm X- \bm Z \rangle  + \frac{\rho}{2}\|\bm X - \bm Z\|_F^2,\\
  &\mathrm{Subject\  \ }\bm X \revise{\succcurlyeq} 0 ,\  \bm Z \in \mathcal{P}_\beta,
\end{align*}
where $\bm V  \in\mathbb{R}^{({n_1+ n_2})\times (n_1+n_2)} $ is the dual variable and $\rho>0$ is a hyper-parameter. 

Then the alternating direction method of multipliers (ADMM)  algorithm  solves this optimization problem by minimizing the augmented Lagrangian with respect to different variables \revise{alternatingly}. More explicitly, at the $(t+1)$-th iteration, the following updates are implemented:
\begin{align*}
  &\bm X^{t+1} = \Pi\{\bm Z^{t} + \rho^{-1}(\bm V^{t} + \mu \bm I)\},\\
  &\bm Z^{(t+1)} = \argmin_{\bm Z \in \mathcal{P}_\beta}\frac{1}{n_1n_2} \|{\bT\circ \ep{\bhW}{1/2} \circ\left(\bY-\bm Z_{12}\right)}\|_{F}^2 + \frac{\rho}{2}\|\bm Z - \bm X^{t+1} - \rho^{-1} \bm V^{t}\|_F^2 = \Phi_{\bm T, \bm Y, \bhW, \beta} \{\bm X^{t+1} + \rho^{-1} \bm V^{t}\},\\
  & \bm V^{t+1} = \bm V^{t} + \tau \rho(\bm X^{t+1} - \bm Z^{t+1}),
\end{align*}
where $\Pi(\cdot)$ is the projection to 
the space \revise{\(\{\bm C \in \mathcal{S}: \bm C \succcurlyeq 0\}\)}, and $\Phi_{\bm T, \bm Y, \bhW, \beta}$ is defined in Definition \ref{def:Phi}. Detailed derivation can be found in \citet{Fang-Liu-Toh18} and \citet{Cai-Zhou16}.

\begin{definition}
  \label{def:Phi}
We use $\bm C(i,j)$ to represent the element on the $i$-th row and $j$-th column of a matrix $\bm C$. For the matrix $\bm C \in \mathbb{R}^{(n_1 + n_2)\times (n_1 + n_2)}$, it can be partitioned into 
\[
\bm C=
\left[
\begin{array}{cc}
\bm C_{11} & \bm C_{12} \\
\bm C_{12}^{\tp} & \bm C_{22}
\end{array}
\right], \bm C_{11} \in \mathbb{R}^{n_1\times n_1}, \bm C_{22} \in \mathbb{R}^{n_2\times n_2}
\]
Then 
\[
\Phi_{\bm T, \bm Y, \bhW, \beta} (\bm C) = 
\left[
\begin{array}{cc}
  \Phi_{\bm T, \bm Y, \bhW, \beta} (\bm C)_{11} & \Phi_{\bm T, \bm Y, \bhW, \beta} (\bm C)_{12} \\
  \Phi_{\bm T, \bm Y, \bhW, \beta}(\bm C)_{12}^{\tp} & \Phi_{\bm T, \bm Y, \bhW, \beta} (\bm C)_{22}
\end{array}
\right],
\]
where
\begin{align*}
  \Phi_{\bm T, \bm Y, \bhW, \beta} (\bm C)_{11}(i,j) &= \min\{\beta, \max\{\bm C_{11}(i,j), -\beta \}\} \qquad  \mathrm{if\ } i\neq j , \\
  \Phi_{\bm T, \bm Y, \bhW, \beta} (\bm C)_{11}(i,j) &= \min\{\beta, \max\{\bm C_{11}(i,j), 0 \}\}\qquad  \mathrm{if\ } i=j,  \\
  \Phi_{\bm T, \bm Y, \bhW, \beta} (\bm C)_{22}(i,j) &=  \min\{\beta, \max\{\bm C_{22}(i,j), -\beta \}\}, \qquad  \mathrm{if\ } i\neq j,  \\
  \Phi_{\bm T, \bm Y, \bhW, \beta} (\bm C)_{22}(i,j) &= \min\{\beta, \max\{\bm C_{22}(i,j), 0 \}\}\qquad   \mathrm{if\ } i=j,  \\
  \Phi_{\bm T, \bm Y, \bhW, \beta} (\bm C)_{12}(i,j) &= \min\left\{\beta, \max\left\{\frac{\bm Y(i,j) \bhW(i,j) + \rho \bm C(i,j)}{\bhW(i,j) + \rho}, -\beta \right\}\right\} \qquad   \mathrm{if\ } \bm T(i,j)= 1 ,\\
  \Phi_{\bm T, \bm Y, \bhW, \beta} (\bm C)_{12}(i,j) &= \min\{\beta, \max\{\bm C_{12}(i,j), -\beta \}\} \  \mathrm{if\ } i\neq j \qquad   \mathrm{if\ } \bm T(i,j)= 0 .\\
\end{align*}
\end{definition}

We summarize the algorithm in Algorithm \ref{alg:convex}. Some piratical implementations to adaptively tune $\rho$ and accelerate the computation can be found in Section 3.3 and 3.4 in \citet{Fang-Liu-Toh18}.

\begin{algorithm}[h]
  \caption{ADMM algorithm}
  \label{alg:convex}
\begin{algorithmic}
  \STATE {\bfseries Input:} $\bY$, $\bT$, $\beta$, $\mu$, $\bhW$,  $\rho = 0.1$, $\tau = 1.618$, K 

  \STATE Initialize $\bm X^{0}$, $\bm Z^{0}$, $\bm V^{0}$, $R$
  \FOR{$t=1$ {\bfseries to} $K-1$}
  \STATE $\bm X^{t+1} \leftarrow \Pi\{\bm Z^{t} + \rho^{-1}(\bm V^{t} + \mu \bm I)\}$
  \STATE $\bm Z^{(t+1)}  \leftarrow \Phi_{\bm T, \bm Y, \bhW, \beta} \{\bm X^{t+1} + \rho^{-1} \bm V^{t}\}$
  \STATE $ \bm V^{t+1}  \leftarrow\bm V^{t} + \tau \rho(\bm X^{t+1} - \bm Z^{t+1})$
  \STATE{Stop if objective value changes less than tolerance}
  \ENDFOR 
\end{algorithmic}
\end{algorithm}

\subsection{Nonconvex Algorithm for Solving \eqref{eqn:objA}}
\label{appsec:algo_nonconvex}
The nonconvex algorithm for max-norm regularization developed in \citet{lee2010practical} base on the equivalent definition  of max-norm  via matrix factorizations:
\begin{align*}
  \|\bm C\|_{\max} := \inf \left\{  \|\bm U \|_{2, \infty}\|\bm V\|_{2, \infty}: \bm C = \bm U \bm V^{\tp}\right\},
\end{align*}
where $\|\cdot\|_{2,\infty}$ denotes the maximum $l_2$ row norm of a matrix. 

To incorporate the nuclear norm regularization, we also notice an equivalent definition of the nuclear norm:
\begin{align*}
  \|\bm C\|_{\revise{\ast}} := \inf \frac{1}{2}\left\{ \|\bm U\|_F^2 + \|\bm V\|_F^2: \bm C = \bm U \bm V^{\tp}\right\}.
\end{align*}

Then we have the following relaxation of the objective function in \eqref{eqn:objA}. 
Take  $$f(\bm L, \bm R) = \frac{1}{n_1n_2} \|{\bT\circ \ep{\bhW}{1/2} \circ\left(\bY-\bm L \bm R^{\tp}\right)}\|_{F}^2 + \frac{\mu}{2} ( \|\bm L\|_F^2 + \|\bm R\|_F^2),$$
and we obtain
\begin{align*}
  & \min_{\bm L, \bm R} f(\bm L, \bm R),\\
  &\mathrm{Subject\  to\ } \max\left\{ \| \bm L\|_{2, \infty}, \| \bm R\|_{2, \infty} \right\} \le \beta.
\end{align*}
\revise{This optimization form is exactly the one in \citet{lee2010practical} except that we add another nuclear penalty in the objective function \(f\). }

\revise{Like what \citet{lee2010practical} considered,} the projected gradient descent method can be applied to iteratively solve this problem. 
We define the project $\mathcal{P}_{B}$ as the Euclidean projection onto the set $\{ \bm M: \|\bm M \|_{2,\infty} \leq B\}$. This projection can be computed by re-scaling the rows of current input matrix whose norms exceed $B$ so their norms equal $B$. Rows with norms less than $B$ are unchanged by the projection. 
We summarize the algorithm in Algorithm \ref{alg:nonconvex}.

\begin{algorithm}[h]
  \caption{Projected gradient descent algorithm 
  }
  \label{alg:nonconvex}
\begin{algorithmic}
  \STATE {\bfseries Input:} $\bY$, $\bT$, $\beta$, $\mu$, $\bhW$,  step size $\tau$, K 
  \STATE Initialize $\bm L^{0}$, $\bm R^{0}$,
  \FOR{$t=1$ {\bfseries to} $K-1$}
  \STATE $\bm L^{t+1} \leftarrow \mathcal{P}_{\beta}\left(\bm L - \tau \frac{\partial f}{\partial \bm L}\right)$
  \STATE $\bm R^{t+1} \leftarrow \mathcal{P}_{\beta}\left(\bm R - \tau \frac{\partial f}{\partial \bm R}\right)$
  \STATE{Stop if objective value changes less than tolerance}
  \ENDFOR 
\end{algorithmic}
\end{algorithm}

\section{Additional Simulation Results}
\label{appsec:sim}
The simulation results for SNR = 1 and SNR = 10 are shown in Table \ref{tab:sim_SNR1} and \ref{tab:sim_SNR10} respectively.

\begin{table}[h]
    \centering
    \small
    \caption{Similar to Table \ref{tab:sim_SNR5}, but for SNR = 1.}
\begin{tabular}{c|ccc}
\hline 
  & \multicolumn{3}{c}{Setting 1}  \tabularnewline
 Method  & $\overline{\mbox{RMSE}}$ & $\overline{\mbox{TE}}$ & \=r \\
 \hline
 \proposed{} & \textbf{1.901(0.004)} & \textbf{1.918(0.004) }& 13.69(0.097)\\ 
 \unitrace & 1.944(0.004) & 1.961(0.004) & 19.55(0.092\\ 
 \unimax & 2.052(0.004) & 2.044(0.004) & 27.695(0.128)  \\ 
 \unihy & \textbf{1.927(0.004)} & \textbf{1.946(0.004)} & 15.265(0.105) \\ 
  \NW & 2.012(0.004) & 2.01(0.004) & 25.61(0.069) \\ 
  \KLT & 2.439(0.005) & 2.492(0.005) & 10.175(0.063)\\ 
  \hline

 & \multicolumn{3}{c}{Setting 2}  \tabularnewline
 Method  & $\overline{\mbox{RMSE}}$ & $\overline{\mbox{TE}}$ & \=r \\
 \hline
\proposed{} & \textbf{1.716(0.004)} & \textbf{1.669(0.004)} & 14.73(0.113) \\ 
  \unitrace & 1.721(0.004) & 1.685(0.004) & 16.335(0.107) \\ 
  \unimax & 1.86(0.004) & 1.799(0.004) & 25.965(0.115) \\ 
  \unihy & \textbf{1.711(0.004)} & \textbf{1.674(0.004)} & 14.565(0.102)  \\ 
  \NW & 1.805(0.005) & 1.747(0.005) & 37.82(0.422) \\ 
  \KLT & 2.16(0.005) & 2.093(0.005) & 2.065(0.110) \\ 
 \hline

  & \multicolumn{3}{c}{Setting 3 } \tabularnewline
 Method  & $\overline{\mbox{RMSE}}$ & $\overline{\mbox{TE}}$ & \=r \\
 \hline
\proposed{}  & \textbf{2.412(0.006) }& \textbf{2.586(0.007)} & 12.495(0.098) \\ 
  \unitrace  & 2.923(0.007) & 3.113(0.007) & 29.15(0.112) \\ 
  \unimax  & 2.641(0.006) & 2.812(0.006) & 28.695(0.109) \\ 
  \unihy & 2.878(0.007) & 3.097(0.007) & 20.105(0.105) \\ 
  \NW  & \textbf{2.668(0.006)} & \textbf{2.779(0.007)} &33.115(0.066) \\ 
  \KLT & 3.667(0.007) & 3.969(0.007) & 9.765(0.067) \\ 
 \hline

\end{tabular}

    \label{tab:sim_SNR1}
\end{table}

\begin{table}[h]
  \centering
  \small
    \caption{Similar to Table \ref{tab:sim_SNR5}, but for SNR = 10.}
\begin{tabular}{c|ccc}
\hline 
& \multicolumn{3}{c}{Setting 1}\tabularnewline
Method  & $\overline{\mbox{RMSE}}$ & $\overline{\mbox{TE}}$ & \=r\\
\hline
\proposed{} & \textbf{0.425(0.001)} & \textbf{0.443(0.001)} & 30.565(0.143) \\ 
  \unitrace & 0.483(0.001) & 0.501(0.001) & 91.615(1.845) \\ 
  \unimax & 0.663(0.001) & 0.688(0.001) & 54.225(0.178) \\ 
  \unihy & 0.427(0.001) & 0.446(0.001) & 32.105(0.129)\\ 
\NW & \textbf{0.405(0.001)}&\textbf{0.422(0.001)}& 30.180(0.196)\\
  \KLT & 1.894(0.003) & 1.958(0.003) & 8.665(0.059) \\ 
\hline

& \multicolumn{3}{c}{Setting 2}  \tabularnewline
Method  & $\overline{\mbox{RMSE}}$ & $\overline{\mbox{TE}}$ & \=r \\
\hline
\proposed{} & \textbf{0.401(0.001)} & \textbf{0.418(0.001)} & 29.360(0.136) \\ 
  \unitrace & 0.475(0.001) & 0.484(0.001) & 120.670(3.387) \\ 
  \unimax & 0.684(0.002) & 0.722(0.002) & 59.300(0.485) \\ 
  \unihy & \textbf{0.409(0.001)} & \textbf{0.426(0.001)} &30.380(0.145)\\ 
  \NW &0.496(0.003)&0.510(0.003)&19.055(0.368)\\
  \KLT & 1.975(0.006) & 1.873(0.004) & 1.375(0.144) \\ 
\hline

& \multicolumn{3}{c}{Setting 3} \tabularnewline
Method &$\overline{\mbox{RMSE}}$ & $\overline{\mbox{TE}}$  &\=r\\
\hline
\proposed{} & \textbf{0.627(0.001)} & \textbf{0.688(0.002) }& 32.675(0.144) \\ 
  \unitrace & 0.868(0.002) & 0.966(0.003) & 77.115(1.343) \\ 
  \unimax & 0.999(0.003) & 1.105(0.004) & 56.890(0.152) \\ 
  \unihy & \textbf{0.670(0.002) }& \textbf{0.736(0.002) }& 39.740(0.820) \\ 
  \NW & 0.703(0.003)&0.768(0.003)&21.860(0.614)\\
  \KLT & 3.157(0.006) & 3.460(0.006) & 9.535(0.088) \\ 
\hline

\end{tabular}
  \label{tab:sim_SNR10}
\end{table}

\bibliographystyle{chicago}
\bibliography{MC_Balancing.bib}

\makeatletter\@input{xx.tex}\makeatother